\documentclass[runningheads]{llncs}
\usepackage{graphicx}
\usepackage{tikz}
\usepackage{comment}
\usepackage{amsmath,amssymb} % define this before the line numbering.
\usepackage{color}
\UseRawInputEncoding

\usepackage[accsupp]{axessibility}  % Improves PDF readability for those with disabilities.
\usepackage{url}
\usepackage{epsfig}
\usepackage{graphicx}
\usepackage{amsmath}
\usepackage{amssymb}
\usepackage{bm}
\usepackage{bbm}
\usepackage{array}
\usepackage{booktabs}
\usepackage{multirow}
\usepackage{makecell}
\usepackage{footmisc}
\usepackage{caption}
\usepackage{subcaption}
\usepackage{footnote}
\usepackage{tablefootnote}
\newcommand{\eat}[1]{}
\usepackage{bbding}

\usepackage[linesnumbered,boxed,ruled,commentsnumbered]{algorithm2e}
\DeclareMathOperator{\clip}{clip}

\DeclareMathOperator{\sign}{sign}

\usepackage[pagebackref,breaklinks,colorlinks]{hyperref}

\usepackage[capitalize]{cleveref}
\crefname{section}{Sec.}{Secs.}
\Crefname{section}{Section}{Sections}
\Crefname{table}{Table}{Tables}
\crefname{table}{Tab.}{Tabs.}

\begin{document}
% \linenumbers
\pagestyle{headings}
\mainmatter
\def\ECCVSubNumber{974}  % Insert your submission number here

\title{Frequency Domain Model Augmentation for Adversarial Attack} % Replace with your title

\begin{comment}
\titlerunning{ECCV-22 submission ID \ECCVSubNumber} 
\authorrunning{ECCV-22 submission ID \ECCVSubNumber} 
\author{Anonymous ECCV submission}
\institute{Paper ID \ECCVSubNumber}
\end{comment}
%******************

% \begin{comment}
\titlerunning{Frequency Domain Model Augmentation for Adversarial Attack}
\author{Yuyang Long\inst{1} 
\and Qilong Zhang\inst{1} 
\and Boheng Zeng\inst{1}
\and Lianli Gao\inst{1} 
\and Xianglong Liu\inst{2} 
\and Jian Zhang\inst{3} 
\and Jingkuan Song\inst{1}\thanks{Corresponding author}}
\authorrunning{Y.Long et al.}
\institute{Center for Future Media, University of Electronic Science and Technology of China \mbox{\and Beihang University \and Hunan University} \\ \email{yuyang.long@outlook.com,\{qilong.zhang,boheng.zeng\}@std.uestc.edu.cn,\\lianli.gao@uestc.edu.cn,xlliu@buaa.edu.cn,jianzh@hnu.edu.cn,\\jingkuan.song@gmail.com}}

% \end{comment}
%******************
\maketitle

\begin{abstract}
For black-box attacks, the gap between the substitute model and the victim model is usually large, which manifests as a weak attack performance.
Motivated by the observation that the transferability of adversarial examples can be improved by attacking diverse models simultaneously, model augmentation methods which simulate different models by using transformed images are proposed. 
However, existing transformations for spatial domain do not translate to significantly diverse augmented models.
To tackle this issue, we propose a novel spectrum simulation attack to craft more transferable adversarial examples against both normally trained and defense models. 
Specifically, we apply a spectrum transformation to the input and thus perform the model augmentation in the frequency domain. We theoretically prove that the transformation derived from frequency domain leads to a diverse spectrum saliency map, an indicator we proposed to reflect the diversity of substitute models. 
Notably, our method can be generally combined with existing attacks. 
Extensive experiments on the ImageNet dataset demonstrate the effectiveness of our method, \textit{e.g.}, attacking nine state-of-the-art defense models with an average success rate of \textbf{95.4\%}.
Our code is available in \url{https://github.com/yuyang-long/SSA}.

\end{abstract}

\section{Introduction}
In recent years, deep neural networks (DNNs) have achieved a considerable success in the field of computer vision, \textit{e.g.}, image classification~\cite{res152,densenet,zhang2022progressive}, face recognition~\cite{deepface,cosface} and self-driving~\cite{bojarski2016end,sallab017deep}. Nevertheless, there are still many concerns regarding the stability of neural networks. As demonstrated in prior works~\cite{Intriguing,fgsm}, adversarial examples which merely add human-imperceptible perturbations on clean images can easily fool state-of-the-art DNNs. Therefore, to help improve the robustness of DNNs, crafting adversarial examples to cover as many blind spots of DNNs as possible is necessary.

\begin{figure*}[h]
    \centering
    \includegraphics[height=6cm]{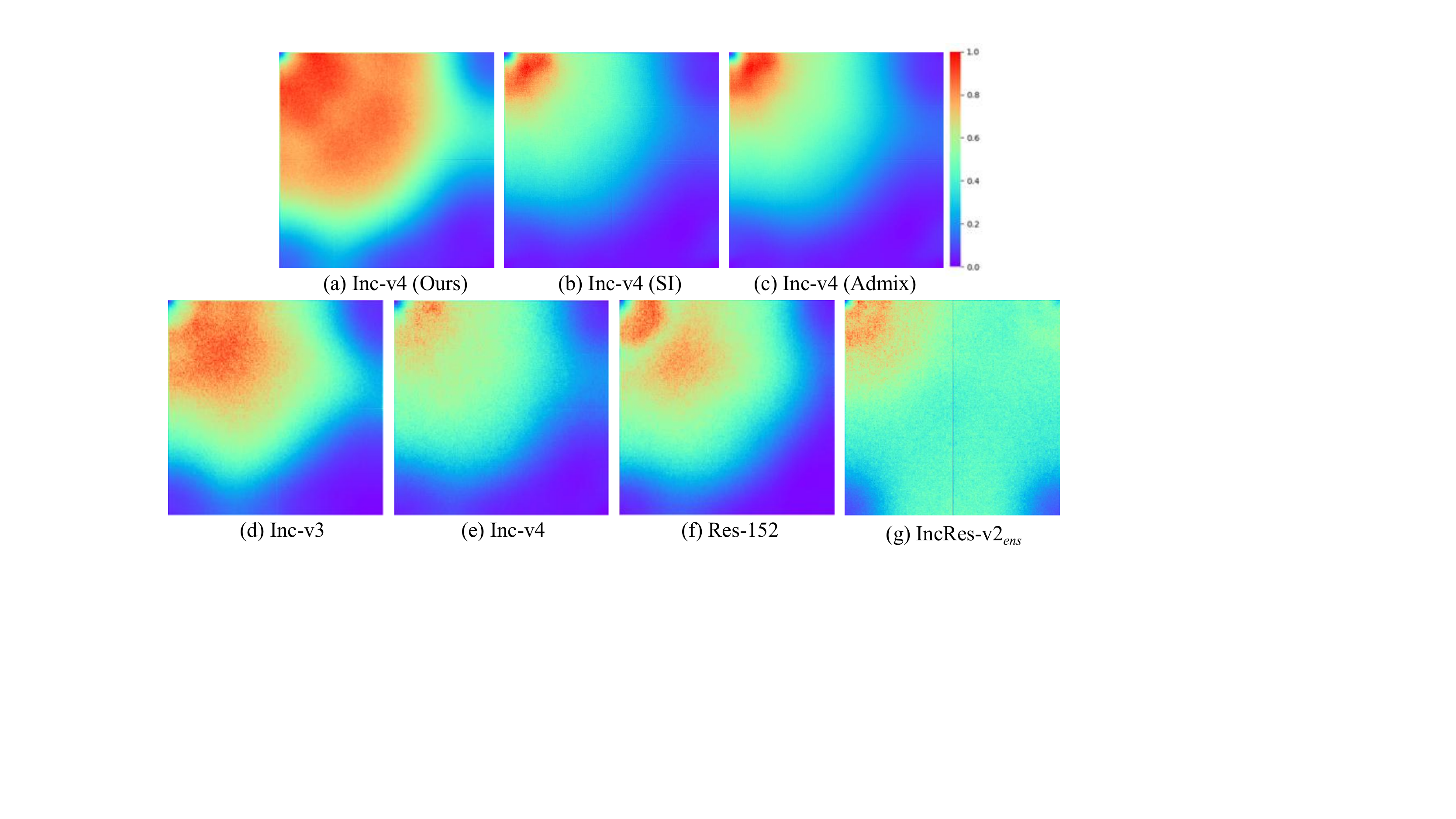}
\caption{Visualization of the spectrum saliency maps (average of all images) for normally trained models Inc-v3~\cite{inc-v3}, Inc-v4~\cite{inc-v4}, Res-152\cite{res152} and defense model IncRes-v2$_{ens}$~\cite{ens}. \textbf{(a)}: the result for our transformation images ($N=5$) conducted in frequency domain. \textbf{(b$\sim$c)}: the result for scale-invariant ($m_1=5$)~\cite{sinifgsm} and Admix ($m_1=5$, $m_2=3$)~\cite{admix} transformations conducted in spatial domain. \textbf{(d$\sim$g)}: the results for raw images on four different models.}
% This clearly reveals that compared with transformation of spatial domain, our generated spectrum saliency map covers almost all other models’ saliency areas, which narrows the gap between models.}
    \label{fig:fre}
\end{figure*}

% In general, adversarial attacks can fall into two main lines: white-box and black-box. For the former~\cite{c&w,deepfool,color,caa}, the attack can get the full knowledge of the victim's model, \textit{e.g.}, the model architecture and parameters. Therefore, the adversarial example can be directly guided by the gradient (w.r.t. the input) of the victim's model, and thus achieving a high success rate. However, the above line is usually impracticable in real-world applications because a model owner is impossible to disclose all information about the deployed model for security purposes. 

In general, settings of adversarial attacks can be divided into white-box and black-box. For the former~\cite{c&w,deepfool,color,caa}, an adversary has access to the model, \textit{e.g.}, model architecture and parameters are known. Therefore, adversarial examples can be directly crafted by the gradient (w.r.t. the input) of the victim model, and thus achieving a high success rate. However, white-box attack is usually impracticable in real-world applications because 
an adversary is impossible to obtain all information about a victim model.
% a model owner is impossible to disclose all information about the deployed model for security purposes. 
To overcome this limitation, a common practice of black-box attacks~\cite{mifgsm,difgsm,pifgsm} turns to investigate the inherent cross-model transferability of adversarial examples. Typically, an adversary crafts adversarial examples via a substitute model (\textit{a.k.a.} white-box model), and then transfers them to a victim model (\textit{a.k.a.} black-box model) for attacking.

% However, the gap between the substitute model and victim's model is usually large, which inevitably increases the difficulty of the attack. To narrow the gap, \textit{model augmentation}~\cite{sinifgsm} is proposed. Typical model augmentation approaches~\cite{difgsm,tifgsm,sinifgsm} aim to simulate different models by applying transformations to the inputs. For instance, at each iteration, Xie \textit{et al.}~\cite{difgsm} apply random resizing and padding to inputs. Dong \textit{et al.}~\cite{tifgsm} shift inputs along the two-dimensions. Lin \textit{et al.}~\cite{sinifgsm} scale inputs with factors $1/2^i$ ($i$ is a hyper-parameter). Yet, these works usually investigate relationships of different models in spatial domain, which might ignore valuable information that is perceptible only in frequency domain.%It is because that there is usually no certain pattern in the representation of images in spatial domain.

However, the gap between the substitute model and the victim model is usually large, which manifests as the low transferability of adversarial examples. Although attacking diverse models simultaneously can boost the transferability, collecting a large number of diverse models is difficult and training a model from scratch is also time-consuming. To tackle this issue, \textit{model augmentation}~\cite{sinifgsm} is proposed. In particular, typical model augmentation approaches~\cite{difgsm,tifgsm,sinifgsm} aim to simulate diverse models by applying loss-preserving transformations to inputs. 
% For instance, at each iteration, 
%Xie \textit{et al.}~\cite{difgsm} apply random resizing and padding to inputs. 
% Dong \textit{et al.}~\cite{tifgsm} shift inputs along two-dimensions. Lin \textit{et al.}~\cite{sinifgsm} scale inputs with factors $1/2^i$ ($i$ is a hyper-parameter). 
Yet, all of existing works investigate relationships of different models in spatial domain, which may overlook the essential differences between them. 
% It is because that different models usually rely on different frequency components of input images when making decisions~\cite{wang2020towards,fourier,highfrequency}. 

% As shown in Figure~\ref{fig:fre} (b \& d), with the scale-invariant~\cite{sinifgsm} transformation in the spatial domain, the spectrum saliency maps (introduced in Sec.~\ref{motivation}) does not change much compared to that of raw image.

To better uncover the differences among models, we introduce the spectrum saliency map (see Sec.~\ref{motivation}) from a frequency domain perspective since representation of images in this domain have a fixed pattern~\cite{dct,fourier}, \textit{e.g.}, low-frequency components of an image correspond to its contour. 
Specifically, the spectrum saliency map is defined as the gradient of model loss function w.r.t. the frequency spectrum of input image. 
As illustrated in Figure~\ref{fig:fre} (d$\sim$g), spectrum saliency maps of different models significantly vary from each other, which clearly reveals that each model has different interests in the same frequency component. 

Motivated by this, we consider tuning the spectrum saliency map to simulate more diverse substitute models, thus generating more transferable adversarial examples. 
To that end, we propose a spectrum transformation based on (discrete cosine transform) DCT and (inverse discrete cosine transform) IDCT techniques~\cite{dct} to diversify input images. 
We theoretically prove that this spectrum transformation can generate diverse spectrum saliency maps and thus simulate diverse substitute models.
% Specifically, with the help of (discrete cosine transform) DCT and (inverse discrete cosine transform) IDCT techniques~\cite{dct}, we propose a spectrum transformation to diversify input images at each iteration. Consequently, elements of the resulting spectrum saliency map can be randomly enlarged or shrank.
%can be mathematically proved to randomly enlarge or shrink elements of the resulting spectrum saliency map. 
%In particular, we apply a spectrum transformation multiple times at each iteration to randomly enlarge or shrink elements of the resulting spectrum, thus to enable more frequency region to be focused on. 
As demonstrated in Figure~\ref{fig:fre} (a$\sim$c), after averaging results of diverse augmented models, only our resulting spectrum saliency map can cover almost all those of other models. This demonstrates our proposed spectrum transformation can effectively narrow the gap between the substitute model and victim model. 
% However, the DCT/IDCT introduced in our spectrum transformation is computationally expensive because input images are high-dimensional and splitting them into blocks before applying DCT/IDCT limits the transferability of adversarial examples.
% To speed up our attack, we further propose a gradient estimation method that uses truncated back-propagated gradient to craft adversarial examples. 
% Therefore, we propose a gradient estimation method to speed up the attack without affecting attack results. Specifically, we truncate the back-propagated gradient at transformation outputs and average gradients of transformation outputs to update adversarial examples.
% To that end, we propose a spectrum transformation to the input image multiple times and average the gradients with respect to each transformation output image to craft adversarial example. The overview of our method is shown in Figure~\ref{fig:process}. Specifically, our attack can be divided into two parts: 1) Spectrum transformation. Add random noise to input image and transform it from spatial domain to frequency domain by the \textit{discrete cosine transform} (DCT)\cite{dct}. Dot product with a scaling matrix to enlarge or shrink the elements of the resulting spectrum randomly. Apply the \textit{inverse discrete cosine transform} (IDCT) to obtain diverse output images in spatial domain. 2) Gradient estimation. Average the gradients of transformation outputs to update adversarial example.
To sum up, our main contributions are as follows:

1) We discover that augmented models derived from the spatial domain transformations are not significantly diverse, which may limit the transferability of adversarial examples. 
% we introduce the spectrum saliency map to uncover the differences among models.

2) To overcome this limitation, we introduce the spectrum saliency map (based on a frequency domain perspective) to investigate the differences among models. Inspired by our finds, we propose a novel Spectrum Simulation Attack to effectively narrow the gap between the substitute model and victim model.

3) Extensive experiments on the ImageNet dataset highlight the effectiveness of our proposed method. Remarkably, compared to state-of-the-art transfer-based attacks,
our method improves the attack success rate by 6.3\%$\sim$12.2\% for normally trained models and 5.6\%$\sim$23.1\% for defense models.

\section{Related Works}
\subsection{Adversarial Attacks}
Since Szegedy \textit{et al.}~\cite{Intriguing} discover the existence of adversarial examples, various attack algorithms~\cite{fgsm,ifgsm,c&w,deepfool,papernot2017practical,uap,color,cda,advdrop,qair,ssm,ATTA,bia,hit,a3} have been proposed to investigate the vulnerability of DNNs. Among all attack branches, FGSM-based black-box attacks~\cite{fgsm,ifgsm,mifgsm,difgsm,pifgsm,pifgsm++,sgm,gao2021feature} which resort to the transferability of adversarial examples are one of the most efficient families. Therefore, in this paper, we mainly focus on this family to boost adversarial attacks.

To enhance the transferability of adversarial examples, it is crucial to avoid getting trapped in a poor local optimum of the substitute model. Towards this end, Dong \textit{et al.}~\cite{mifgsm} adopt the momentum term at each iteration of I-FGSM~\cite{ifgsm} to stabilize update direction. Lin \textit{et al.}~\cite{sinifgsm} further adapt Nesterov accelerated gradient~\cite{Nesterov1983AMF} into the iterative attacks with the aim of effectively looking ahead.
Gao \textit{et al.}~\cite{pifgsm} propose patch-wise perturbations to better cover the discriminate region of images. 
In addition to considering better optimization algorithms, \textit{model augmentation}~\cite{sinifgsm} is also an effective strategy. Xie \textit{et al.}~\cite{difgsm} introduce a random transformation to the input, thus improving the transferability. Dong \textit{et al.}~\cite{tifgsm} shift the input to create a series of translated images and approximately estimate the overall gradient to mitigate the problem of over-reliance on the substitute model. Lin \textit{et al.}~\cite{sinifgsm} leverage the scale-invariant property of DNNs and thus average the gradients with respect to different scaled images to update adversarial examples. Zou \textit{et al.}~\cite{dem} modify DI-FGSM~\cite{difgsm} to promote TI-FGSM~\cite{tifgsm} by generating multi-scale gradients. Wang \textit{et al.}~\cite{vtfgsm} consider the gradient variance along momentum optimization path to avoid overfitting. Wang \textit{et al.}~\cite{fia} average the gradients with respect to feature maps to disrupt important object-aware features. Wang \textit{et al.}~\cite{admix} average the gradients of a set of admixed images, which are the input image admixed with a small portion of other images while maintaining the label of the input image. Wu \textit{et al.}~\cite{ATTA} utilizes an adversarial transformation network to find a better transformation for adversarial attacks in the spatial domain.
\subsection{Frequency-based Analysis and Attacks}
Several works~\cite{fourier,highfrequency,EffectivenessofLow,advdrop,wang2020towards} have analyzed DNNs from a frequency domain perspective. Wang \textit{et al.}~\cite{highfrequency} notice DNNs' ability in capturing high-frequency components of an image which are almost imperceptible to humans. Dong \textit{et al.}~\cite{fourier}  find that naturally trained models are highly sensitive to additive perturbations in high frequencies, and both Gaussian data augmentation and adversarial training can significantly improve robustness against high-frequency noises.
% naturally trained model is highly sensitive to additive noise in all but the lowest frequencies and adversarial training dramatically improve robustness in the higher frequencies but sacrificing the robustness of the naturally trained model in the lowest frequencies.

In addition, there also exists several adversarial attacks based on frequency domain. Guo \textit{et al.}~\cite{LF} propose a LF attack that only leverages the low-frequency components of an image, which shows that low-frequency components also play a significant role in model prediction as high-frequency components.
% does utilize the features in the low-frequency domains for predictions instead of only learning from high-frequency components.
Sharma \textit{et al.}~\cite{EffectivenessofLow} demonstrate that defense models based on adversarial training are less sensitive to high-frequency perturbations but still vulnerable to low-frequency perturbations. Duan \textit{et al.}~\cite{advdrop} propose the AdvDrop attack which generates adversarial examples by dropping existing details of clean images in frequency domain. Unlike these works that perturb a subset of frequency components, our method aims to narrow the gap between models by frequency-based analysis.

\subsection{Adversarial Defenses}
% To mitigate the threat of adversarial examples, numerous adversarial defense techniques~\cite{fgsm,madry2018towards,NRP,Guo2018Countering} have been proposed in recent years. 
% Tram{\`{e}}r \textit{et al.}~\cite{ens} introduce ensemble adversarial training, which augments training data with perturbation from other models, to improve model robustness further. 
% Xie \textit{et al.}~\cite{xie2019feautre} inject blocks that can denoise the intermediate features into the network, and then end-to-end train it with adversarial examples to reduce perturbations in feature maps. 
% However, adversarial training is very time-consuming. Hence many works try to purify the adversarial examples before feeding to DNNs. ~\cite{Guo2018Countering,RP} utilize multiple input transformations to mitigate adversarial effects. Liao \textit{et al.}~\cite{HGD} propose high-level representation guided denoiser to suppress the influence of adversarial perturbation. Cohen \textit{et al.}~\cite{RS} leverage the classifier with Gaussian data augmentation to enhance model robustness. Naseer \textit{et al.}~\cite{NRP} design a Neural Representation Purifier (NRP) model that learns to clean adversarial perturbed images based on the automatically derived supervision.

To mitigate the threat of adversarial examples, numerous adversarial defense techniques have been proposed in recent years. One popular and promising way is adversarial training~\cite{fgsm,madry2018towards} which leverages adversarial examples to augment the training data during the training phase. Tram{\`{e}}r \textit{et al.}~\cite{ens} 
introduce ensemble adversarial training, which decouples the generation of adversarial examples from the model being trained, to yield models with stronger robustness to black-box attacks. Xie \textit{et al.}~\cite{xie2019feautre} inject blocks that can denoise the intermediate features into the network, and then end-to-end train it on adversarial examples to learn to reduce perturbations in feature maps. 

Although adversarial training is the most effective strategy to improve the robustness of the model at present, it inevitably suffers from time-consuming training costs and is expensive to be applied to
large-scale datasets and complex DNNs. To avoid this issue, many works try to cure the infection of adversarial perturbations before feeding to DNNs. Guo \textit{et al.}~\cite{Guo2018Countering} utilize multiple input transformations (\textit{e.g.}, JPEG compression~\cite{jpeg}, total variance minimization~\cite{tvm} and image quilting~\cite{imagequilt}) to recover from the adversarial perturbations. Liao \textit{et al.}~\cite{HGD} propose high-level representation guided denoiser (HGD) to suppress the influence of adversarial perturbation. Xie \textit{et al.}~\cite{RP} mitigate adversarial effects through random resizing and padding (R\&P). Cohen \textit{et al.}~\cite{RS} leverage the classifier with Gaussian data augmentation to create a provably robust classifier.

In addition, researchers also try to combine the benefits of adversarial training and input pre-processing methods to further improve the robustness of DNNs.
NeurIPS-r3 solution~\cite{nips_r3} propose a two-step procedure which first process images with a series of transformations (\textit{e.g.}, rotation, zoom and sheer) and then pass the outputs through an ensemble of adversarially trained models to obtain the overall prediction. 
Naseer \textit{et al.}~\cite{NRP} design a Neural Representation Purifier (NRP) model that learns to clean adversarial perturbed images based on the automatically derived supervision.

\section{Methodology}
% \TD{In the section,  we first provide details of several adversarial attacks for enhancing the transferability to which our method is most related, and then give an introduction of our motivation. At last, we explicitly illustrate our method and explain its superiority.}

% In this section, we provide the detailed description of our method. In section~\ref{prem}, we give the definition of our task.

In this section, we first give the basic definition of our task in Sec.~\ref{prem}, and then introduce our motivation in Sec.~\ref{motivation}. Based on the motivation, we provide a detailed description of the proposed method - Spectrum Transformation (Sec.~\ref{Spectrum Transformation}). Finally,  we introduce our overall attack algorithm in Sec.~\ref{attal}.
%,in combination with I-FGSM~\cite{ifgsm}.

\subsection{Preliminaries}
\label{prem}
Formally, let $f_\theta:\bm{x}\rightarrow y$ denotes a classification model, where $\theta$, $\bm{x}$ and $y$ indicate the parameters of the model, input clean image and true label, respectively. Our goal is to craft an adversarial perturbation $\bm{\delta}$ so that the resulting adversarial example $\bm{x'}=\bm{x}+\bm{\delta}$ can successfully mislead the classifier, \textit{i.e.}, $f_\theta(\bm{x'})\neq y$ (\textit{a.k.a.} non-targeted attack). To ensure an input is minimally changed, an adversarial example should be in the $\ell_p$-norm ball centered at $\bm{x}$ with radius $\epsilon$. Following previous works~\cite{mifgsm,difgsm,tifgsm,pifgsm,admix,fia,gao2021feature}, we focus on the $\ell_{\infty}$-norm in this paper. Therefore, the generation of adversarial examples can be formulated as the following optimization problem:
\begin{equation}
\label{aim}
% \begin{split}
    \mathop{\arg\max}\limits_{\bm{x'}} J(\bm{x'},y;\theta),\,\,\,\,\text{ s.t.\ $\| \bm{\delta} \|_{\infty}\leq\epsilon$},
% \end{split}
\end{equation}
where $J(\bm{x'},y;\theta)$ is usually the cross-entropy loss. However, it is impractical to directly optimize Eq.~\ref{aim} via the victim model $f_\theta$ under the black-box manner because its parameter $\theta$ is inaccessible. To overcome this limitation, a common practice is to craft adversarial examples via the accessible substitute model $f_\phi$ and relying on the transferability to fool the victim model. Taking I-FGSM~\cite{ifgsm} as an example, adversarial examples at iteration $t+1$ can be expressed as:
\begin{equation}
\label{ifgsm}
    \bm{x'}_{t+1} = \clip_{\bm{x}, \epsilon}\{\bm{x'}_{t} + \alpha \cdot \sign\left(\nabla_{\bm{x'}_t} J\left(\bm{x'}_t, y;\phi\right)\right)\},
\end{equation}
where $\clip_{\bm{x}, \epsilon}(\cdot)$ denotes an element-wise clipping operation to ensure $\bm{x'}\in[\bm{x}-\epsilon, \bm{x}+\epsilon]$, and $\alpha$ is the step size.

% We define $f$ as image classifier with parameter $\theta$ and $x$ as the input image with its true label $y^{true}$. Let $J(\bm{x},y^{true};\theta)$ denote the loss function of classifier $f$ ($i.e.$ the cross-entropy loss). Our goal is to find an adversarial example \bm{$x_{adv}$} that satisfies: 
% In white-box attack, we can transfer this function to an optimation problem, the expression is:
% \begin{equation}
% % \begin{split}
%     \bm{x_{adv}} = \mathop{\arg\max}\limits_{\|\bm{x_{adv}} - \bm{x} \|_{\infty} < \epsilon} J(\bm{x_{adv}},y^{true};\theta)
% % \end{split}
% \end{equation}

\subsection{Spectrum Saliency Map}
\label{motivation}
In order to effectively narrow the gap between models, it is important to uncover the essential differences between them. Recently, various attack methods~\cite{fgsm,ifgsm,mifgsm,difgsm,tifgsm,pifgsm,dem,sinifgsm,vtfgsm,admix} have been proposed to boost the transferability of adversarial examples. Among these algorithms, \textit{model augmentation}~\cite{sinifgsm} is one of the most effective strategies. However, existing works (\textit{e.g.},~\cite{tifgsm,sinifgsm}) usually augment substitute models by applying loss-preserving transformations in the spatial domain, which might ignore essential differences among models and reduce the diversity of substitute models. 
Intuitively, different models usually focus on similar \textit{spatial regions} of each input image since location of key objects in images is fixed.
% it is impossible that different substitute models focus on different \textit{spatial regions} of all input images since models always focus on the key objects in images.
% % they have different layout. 
By contrast, as demonstrated in previous work~\cite{wang2020towards,fourier,highfrequency}, different models usually rely on different \textit{frequency components} of each input image when making decisions. 

% However, existing works~\cite{difgsm,tifgsm,pifgsm,dem,sinifgsm,admix} are usually motivated by the observation in spatial domain, which may 

% At present, \textit{model augmentation}~\cite{sinifgsm} is one of the most effective strategies

% However, previous works~\cite{difgsm,tifgsm,pifgsm,dem,sinifgsm,admix}

% For generating more transferable adversarial examples, various attack methods~\cite{fgsm,ifgsm,mifgsm,difgsm,tifgsm,pifgsm,dem,sinifgsm,vtfgsm,admix} have been proposed. Among these algorithms, \textit{model augmentation}~\cite{sinifgsm} is one of the most effective strategies, and existing works~\cite{tifgsm,sinifgsm,dem} usually apply transformations in the spatial domain. 
% However, as demonstrated in previous work~\cite{wang2020towards}, different models rely on different frequency components of an input image when making decisions. 
% Therefore, model augmentation based on spatial domain might ignore essential differences among models, thus reducing the diversity of substitute models.

Motivated by this, we turn to explore correlations among models from a perspective of frequency domain. Specifically, we adopt DCT to transform input images $\bm{x}$ from the spatial domain to the frequency domain. The mathematical definition of the DCT (denoted as $\mathcal{D}(\cdot)$\footnote{In the implementation, DCT is applied to each color channel independently.} in the following) can be simplified as:
\begin{equation}
    \mathcal{D}(\bm{x}) = \bm{A}\bm{x}\bm{A}^\mathrm{T},
\end{equation} 
where $\bm{A}$ is an orthogonal matrix and thus $\bm{A}\bm{A}^\mathrm{T}$ is equal to the identity matrix $\bm{I}$. Formally, low-frequency components whose amplitudes are high tend to be concentrated in the upper left corner of a spectrum, and high-frequency components are located in the remaining area. Obviously, the pattern of frequency domain is more fixed compared with diverse representations of images in spatial domain (more visualizations can be found in supplementary Sec. D.1).
Therefore, we propose a spectrum saliency map $\bm{S}_\phi$ to mine sensitive points of different models $f_\phi$, which is defined as:
%More concretely, $\bm{S}_\phi$ is defined as:
\begin{equation}
   \bm{S}_\phi = \frac{\partial {J(\mathcal{D_I}(\mathcal{D}(\bm{x})),y;\phi)}}{\partial \mathcal{D}(\bm{x})},\\
\end{equation}
where $\mathcal{D_I}(\cdot)$ denotes the IDCT which can recover the input image from frequency domain back to spatial domain. Note that both the DCT and the IDCT are lossless, \textit{i.e.}, $\mathcal{D_I}(\mathcal{D}(\bm{x}))=A^{\mathrm{T}}\mathcal{D}(\bm{x})A=\bm{x}$.

From the visualization result of $\bm{S}_\phi$ shown in Figure~\ref{fig:fre}, we observe that frequency components of interest usually varies from model to model. Hence, the spectrum saliency map can serve as an indicator to reflect a specific model.
% \TD{
% 
% Besides, previously proposed transformations in the spatial domain (\textit{i.e.}, Figure~\ref{fig:fre}(b~\&~c)) is less effective for generating diverse spectrum saliency maps, leading to a weaker model augmentation. }
%the spectrum saliency maps vary from model to model, 
%which clearly reveals that different models have different interests in the same frequency component.

% different model has different sensitivity for a certain frequency point of a image. Especially the defense model, almost utilize all frequency components besides high-frequency. This motivate us to consider tune the value of frequency domain to fit other models when attack the surrogate model. Figure \ref{fig:process} overviews the proposed Spectrum Simulation (FT) attack (detailed in Section 3.3), which can effectively trade off the difference between surrogate model and target model. Ultimately generate more transferable adversarial examples.
% \subsection{Frequency Heatmap}

\subsection{Spectrum Transformation}
\label{Spectrum Transformation}
The analysis above motivates us that if we can simulate augmented models with a similar spectrum saliency map to victim model, the gap between the substitute model and victim model can be significantly narrowed and adversarial examples can be more transferable.

% Fortunately, we prove that using matrix transformation in the frequency domain can lead to a diverse spectrum saliency map.

% In this section, we turn to enhance the transferability of adversarial examples from a perspective of frequency domain model augmentation.
% Specifically, we consider a spectrum transformation to make spectrum saliency map fit different models.
% Specifically, we propose a spectrum transformation $\mathcal{T}(\cdot)$ that can make the resulting spectrum saliency maps of the substitute model similar to that of the victim's model.

\noindent{\bfseries{Lemma 1}}. \textit{Assume both $\bm{B}_1$ and $\bm{B}_2$ are n-by-n matrix and $\bm{B}_1$ is invertible, then there must exist an n-by-n matrix $\bm{C}$ that can make $\bm{B}_1\times C=\bm{B}_2$.}

Lemma 1 shows that it is possible to make two matrices (note the essence of spectrum saliency map is also a matrix) equal in the form of a matrix transformation. However, the spectrum saliency map of vicitm model is usually not available under black-box setting. Moreover, spectrum saliency map of substitute model is high-dimensional and not guaranteed to be invertible. 
To tackle this problem, we propose a random spectrum transformation $\mathcal{T}(\cdot)$ which decomposes matrix multiplication into matrix addition and Hadamard product to get diverse spectrums. 
Specifically, in combination with the DCT/IDCT, our $\mathcal{T}(\cdot)$ can be expressed as:
\begin{align}
    \mathcal{T}(\bm{x}) &= \mathcal{D_I}((\mathcal{D}(\bm{x})+\mathcal{D}(\bm{\xi})) \odot \bm{M}), \\
    & = \mathcal{D_I}(\mathcal{D}(\bm{x}+\bm{\xi}) \odot \bm{M})
    \label{transform}
\end{align}
where $\odot$ denotes Hadamard product, $\xi\sim \mathcal{N}(0, \sigma^2I)$ and each element of $M\sim \mathcal{U}(1 - \rho, 1 + \rho)$ are random variants sampled from Gaussian distribution and Uniform distribution, respectively. In practice, common DCT/IDCT paradigm~\cite{advdrop,projection}, \textit{i.e.}, splitting the image into several blocks before applying DCT, not works well for boosting transferability (see the ablation study for experimental details). Therefore, we apply DCT on the whole image in our experiments and visualization of transformation outputs can be found in supplementary Sec. D.2.

Formally, $\mathcal{T}(\cdot)$ is capable of yielding diverse spectrum saliency maps (we also provide proof in supplementary Sec. A) which can reflect the diversity of substitute models, and meanwhile, narrowing the gap with the victim model.
As illustrated in Figure~\ref{fig:fre}, previously proposed transformations~\cite{sinifgsm,admix} in the spatial domain (\textit{i.e.}, (b~\&~c)) is less effective for generating diverse spectrum saliency maps, which may lead to a weaker model augmentation. In contrast, with our proposed spectrum transformation, resulting spectrum saliency map (\textit{i.e.}, (a)) can cover almost all those of other models.

\begin{algorithm}[h]
\caption{S$^2$I-FGSM}%算法名字
\label{alg:Framwork} 
\SetAlgoNoLine % 不要算法中的竖线
\SetKwInOut{Input}{\textbf{Input}}\SetKwInOut{Output}{\textbf{Output}}
\LinesNumbered %要求显示行号
\Input{A classifier $f$ with parameters $\phi$; loss function $J$; a clean image \bm{$x$} with ground-truth label $y$; iterations $T$; $L_\infty$ constraint $\epsilon$; spectrum transformation times $N$; tunning factor $\rho$; std $\sigma$ of noise \bm{$\xi$}.}
\Output{The adversarial example \bm{$x'$}}
$\alpha = \epsilon / T$, $\bm{x'_0} = \bm{x}$\\
\For{$t = 0 \rightarrow T-1$}{
    \For{$i = 1 \rightarrow N$}{
    Get transformation output $\mathcal{T}(\bm{x'_{t}})$ using Eq.~\ref{transform}\\
    Gradient calculate $\bm{g'_i}=\nabla_{\bm{x'_{t}}} J(\mathcal{T}(\bm{x'_{t}}), y; \phi)$ 
    }
　  Average gradient: $\bm{g'} = \frac{1}{N}\sum_{i=1}^{N}\bm{g'_i}$\\
　　$\bm{x'_{t+1}} = \clip_{\bm{x},\epsilon}\left\{\bm{x'_{t}} + \alpha \cdot sign(\bm{g'})\right\}$\\
　　$\bm{x'_{t+1}} = \clip(\bm{x'_{t+1}}, 0, 1)$\\
}
\bm${x'}$ = \bm{$x'_T$}\\
\Return \bm${x'}$
\end{algorithm}

\begin{figure*}[h]
    \centering
    \includegraphics[height=7.5cm]{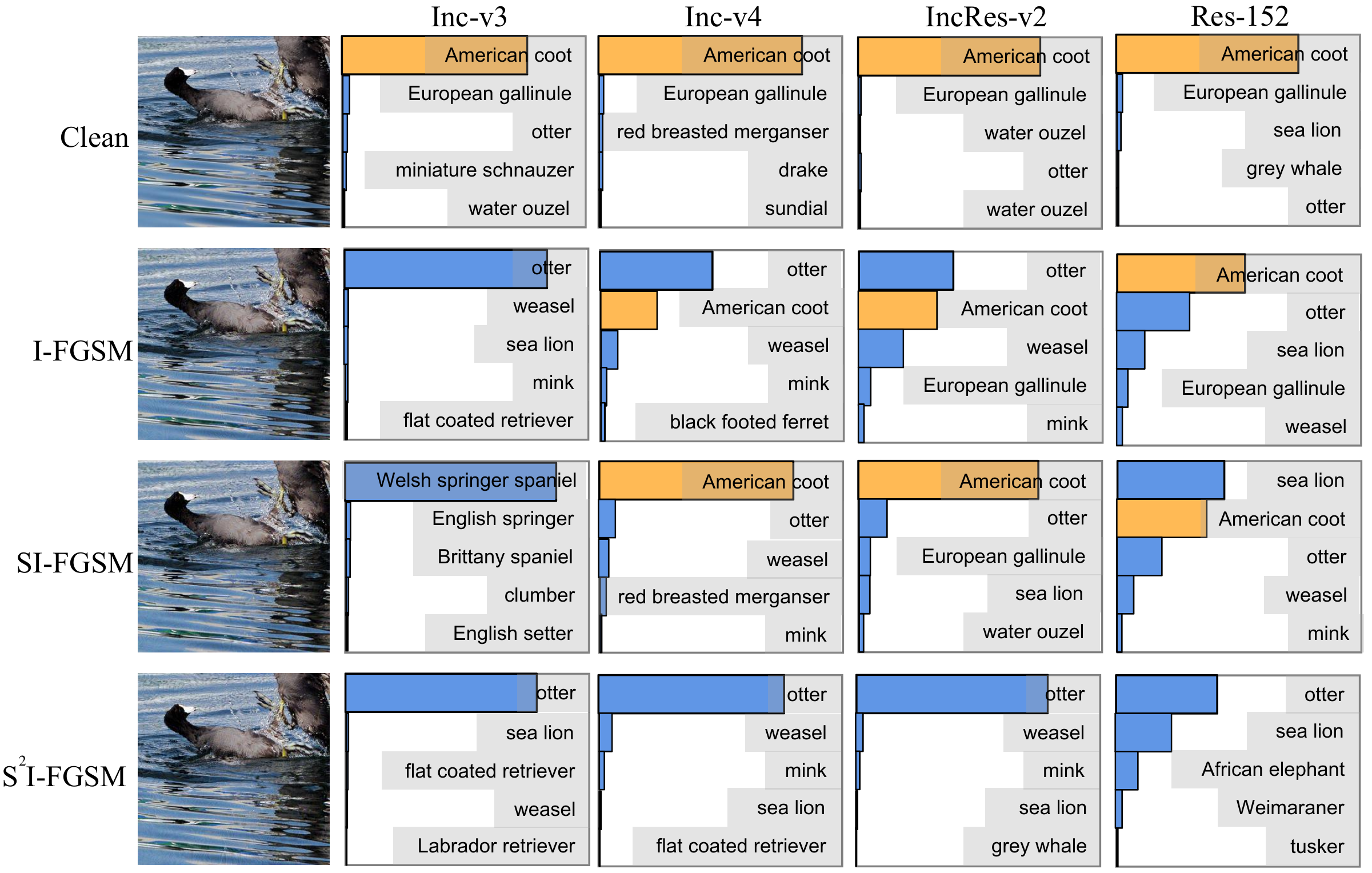}
    \caption{The comparisons of attacks on Inc-v3~\cite{inc-v3}, Inc-v4~\cite{inc-v4}, IncRes-v2~\cite{inc-v4} and Res152~\cite{res152}. The true label of clean image is \textit{American coot} and marked as orange in the top-5 confidence distribution plots.
    The adversarial examples are crafted via Inc-v3~\cite{inc-v3} by I-FGSM~\cite{ifgsm}, SI-FGSM~\cite{sinifgsm} and our proposed S$^2$I-FGSM, respectively. Remarkably, our method can attack the white-box model and all black-box models successfully.}
    \label{si_ft}
\end{figure*}

\subsection{Attack Algorithm}
\label{attal}
% In Sec.~\ref{Spectrum Transformation} and Sec.~\ref{Gradient Estimation}, we only introduce our spectrum transformation and gradient estimation, but do not specify the attack algorithm for crafting adversarial examples. It is because our method can be generally integrated into many gradient-based attacks. For instance, in combination with I-FGSM~\cite{ifgsm} (\textit{i.e.}, Eq.~\ref{ifgsm}), Figure~\ref{fig:process} overviews the proposed Spectrum Simulation Fast Gradient Sign Method (S$^2$I-FGSM).
In Sec.~\ref{Spectrum Transformation}, we have introduced our proposed spectrum transformation. This method could be integrated with any gradient-based attacks. For instance, in combination with I-FGSM~\cite{ifgsm} (\textit{i.e.}, Eq.~\ref{ifgsm}), we propose the Spectrum Simulation Iterative Fast Gradient Sign Method (S$^2$I-FGSM). The algorithm is detailed in Algorithm~\ref{alg:Framwork}. Technically, our attack can be mainly divided into three steps. First, in lines 3-6, we apply our spectrum transformation $\mathcal{T}(\cdot)$ to the input image $\bm{x'_{t}}$ so that the gradient $\bm{g_i'}$ obtained from the substitute model is approximately equal to the result obtained from a 
new model, \textit{i.e.}, \textit{model augmentation}. Second, in line 7, we average $N$ augmented models' gradients to obtain a more stable update direction $\bm{g'}$. Finally, in line 8, we update adversarial examples $\bm{x'_{t+1}}$ of iteration $t+1$. In short, the above process can be summarised in the following formula: 
\begin{equation}
\label{ftfgsm}
\begin{aligned}
    \bm{x'_{t+1}} = \clip_{\bm{x},\epsilon}\{\bm{x'_{t}} + \alpha \cdot sign(\frac{1}{N}\sum_{i=1}^{N} \nabla_{\bm{x'_{t}}} J(\mathcal{T}(\bm{x'_{t}}), y; \phi))\}.
\end{aligned}
\end{equation}
% add the $t$-th perturbation on 
% we apply spectrum transformation $\mathcal{T}$ to input image $\bm{x'_{t}}$ so that the gradient $\bm{g_i'}$ derived from the substitute model is 
% to simulate 
% the update direction of our proposed Spectrum Simulation Fast Gradient Sign Method (S$^2$I-FGSM) can be expressed as:
% \begin{equation}
% \label{ftfgsm}
% \begin{aligned}
%       \bm{x'_{t+1}} =& \clip_{\bm{x}, \epsilon}\{\bm{x'_t} + \alpha\cdot sign(\frac{1}{N}\sum_{i=1}^{N} \nabla_{\bm{x'_{t}}} J(\mathcal{T}(\bm{x'_{t}}), y; \phi))\},
% \end{aligned}
% \end{equation}
% where $N$ indicates the number of augmented models. 
The resulting adversarial examples are shown in Figure~\ref{si_ft}. Compared with I-FGSM~\cite{ifgsm} and SI-FGSM~\cite{sinifgsm}, our proposed S$^2$I-FGSM can craft more threatening adversarial examples for fooling black-box models.

\section{Experiments}
\subsection{Experiment Setup}
\label{setup}
 \hspace{1.5em}{\bfseries{Dataset.}} Following previous works \cite{mifgsm,tifgsm,gao2021feature,pifgsm}, we conduct our experiments on the ImageNet-compatible dataset\footnote{\url{https://github.com/cleverhans-lab/cleverhans/tree/master/cleverhans_v3.1.0/examples/nips17_adversarial_competition/dataset}}, which contains 1000 images with resolution $299\times299\times3$.

{\bfseries{Models.}} We choose six popular normally trained models, including Inception-v3 (Inc-v3) \cite{inc-v3}, Inception-v4 (Inc-v4) \cite{inc-v4}, Inception-Resnet-v2 (IncRes-v2) \cite{inc-v4}, Resnet-v2-50 (Res-50), Resnet-v2-101 (Res-101) and Resnet-v2-152 (Res-152) \cite{res152}. For defenses, we consider nine defense models (\textit{i.e.}, Inc-v3$_{ens3}$, Inc-v3$_{ens4}$, IncRes-v2$_{ens}$ \cite{ens}, HGD \cite{HGD}, R\&P \cite{RP}, NIPS-r3 \cite{nips_r3}, JPEG~\cite{Guo2018Countering}, RS \cite{RS} and NRP \cite{NRP}) that are robust against black-box attacks.

{\bfseries{Competitor.}} To show the effectiveness of our proposed spectrum simulation attack, we compare it with diverse state-of-the-art attack methods, including MI-FGSM~\cite{mifgsm}, DI-FGSM~\cite{difgsm}, TI-FGSM~\cite{tifgsm}, PI-FGSM~\cite{pifgsm}, SI-NI-FGSM~\cite{sinifgsm}, VT-FGSM~\cite{vtfgsm}, FI-FGSM~\cite{fia} and Admix~\cite{admix}. Besides, we also compare the combined version of these methods, \textit{e.g.}, TI-DIM (combined version of TI-FGSM, MI-FGSM and DI-FGSM) and SI-NI-TI-DIM. 

{\bfseries{Parameter Settings.}} In all experiments, the maximum perturbation $\epsilon = 16$, the iteration $T = 10$, and the step size $\alpha = \epsilon/T = 1.6$. For MI-FGSM, we set the decay factor $\mu$ = 1.0. For DI-FGSM, we set the transformation probability $p=0.5$. For TI-FGSM, we set the kernel length $k$ = 7. For PI-FGSM, we set the amplification factor $\beta$ = 10, project factor $\gamma$ = 16 and the kernel length $k_w$ = 3 for normally trained models, $k_w$ = 7 for defense models. For SI-NI-FGSM, we set the number of copies $m_1$ = 5. For VT-FGSM, we set the hyper-parameter $\beta$ = 1.5, number of sampling examples is 20. For FI-FGSM, the drop probability $p_d$ = 0.3 for normally trained models and $p_d$ = 0.1 for defense models, and the ensemble number is 30. For Admix, we set number of copies $m_1=5$\footnote{Note that Admix is equipped with SI-FGSM by default.} , sample number $m_2$ = 3 and the admix ratio $\eta$ = 0.2. For our proposed S$^2$I-FGSM, we set the tuning factor $\rho$ = 0.5 for $\bm{M}$, the standard deviation $\sigma$ of $\bm{\xi}$ is simply set to the value of $\epsilon$, and the number of spectrum transformations $N$ = 20 (discussions about $\rho$, $\sigma$ and $N$ can be found in supplementary Sec. B). The parameter settings for the combined version are the same.

\begin{table*}[t]
\centering
\caption{The attack success rates (\%) on six normally trained models. The adversarial examples are crafted via Inc-v3, Inc-v4, IncRes-v2 and Res-152, respectively. ``*" indicates white-box attacks.}
\resizebox{0.85\linewidth}{!}{
\begin{tabular}{c|c|p{1.4cm}<{\centering}|p{1.4cm}<{\centering}|c|p{1.4cm}<{\centering}|c|c|c}
\hline
Model & Attack & Inc-v3 & Inc-v4 & IncRes-v2 & Res-152 & Res-50 & Res-101 & AVG. \\
\hline
\hline
\multirow{9}{*}{Inc-v3} & MI-FGSM & \textbf{100.0*} & 50.6 & 47.2 & 40.6 & 46.9 & 41.7 & 54.5 \\
 & DI-FGSM & 99.7* & 48.3 & 38.2 & 31.8 & 39.0 & 33.8 & 48.5 \\
 & PI-FGSM & \textbf{100.0*} & 56.5 & 49.6 & 45.0 & 50.1 & 44.7 & 57.7 \\
 & S$^2$I-FGSM(ours) & 99.7* & \textbf{65.0} & \textbf{58.9} & \textbf{50.3} & \textbf{56.2} & \textbf{53.3} & \textbf{63.9} \\
 \cline{2-9}
 & SI-NI-FGSM & \textbf{100.0*} & 76.0 & 75.8 & 67.7 & 73.0 & 69.4 & 77.0 \\
 & VT-MI-FGSM & \textbf{100.0*} & 75.0 & 69.6 & 62.7 & 67.1 & 63.1 & 72.9 \\
 & FI-MI-FGSM & 98.8* & 83.6 & 80.0 & 72.7 & 80.2 & 74.9 & 81.7 \\
 & S$^2$I-MI-FGSM(ours) & 99.6* & \textbf{88.2} & \textbf{85.8} & \textbf{81.0} & \textbf{83.4} & \textbf{81.3} & \textbf{86.6} \\
 \hline
\multirow{9}{*}{Inc-v4} & MI-FGSM & 62.0 & \textbf{100.0*} & 46.2 & 41.4 & 47.7 & 42.8 & 56.7 \\
 & DI-FGSM & 54.1 & 99.1* & 36.3 & 31.4 & 33.7 & 30.4 & 47.5 \\
 & PI-FGSM & 60.3 & \textbf{100.0*} & 45.9 & 44.1 & 50.3 & 42.7 & 57.2 \\
 & S$^2$I-FGSM(ours) & \textbf{70.2} & 99.6* & \textbf{57.1} & \textbf{48.1} & \textbf{56.5} & \textbf{47.7} & \textbf{63.2} \\
 \cline{2-9}
 & SI-NI-FGSM & 83.8 & \textbf{99.9*} & 78.2 & 73.3 & 77.0 & 73.9 & 81.0 \\
 & VT-MI-FGSM & 77.8 & 99.8* & 71.5 & 64.1 & 65.7 & 64.4 & 73.9 \\
 & FI-MI-FGSM & 84.9 & 94.7* & 78.0 & 75.4 & 78.0 & 75.7 & 81.1 \\
 & S$^2$I-MI-FGSM(ours) & \textbf{90.3} & 99.6* & \textbf{86.5} & \textbf{83.1} & \textbf{83.3} & \textbf{81.0} & \textbf{87.3} \\
 \hline
\multirow{9}{*}{IncRes-v2} & MI-FGSM & 60.4 & 52.8 & 99.4* & 45.9 & 49.1 & 46.3 & 59.0 \\
 & DI-FGSM & 56.5 & 49.1 & 97.8* & 35.6 & 38.3 & 37.1 & 52.4 \\
 & PI-FGSM & 62.6 & 57.9 & \textbf{99.5*} & 47.0 & 51.4 & 47.9 & 61.1 \\
 & S$^2$I-FGSM(ours) & \textbf{76.0} & \textbf{67.7} & 98.3* & \textbf{56.2} & \textbf{59.8} & \textbf{58.4} & \textbf{69.4} \\
 \cline{2-9}
 & SI-NI-FGSM & 86.4 & 82.3 & \textbf{99.8*} & 76.8 & 79.6 & 76.4 & 83.4 \\
 & VT-MI-FGSM & 79.3 & 75.6 & 99.5* & 66.8 & 69.5 & 69.5 & 76.7 \\
 & FI-MI-FGSM & 81.9 & 77.9 & 89.2* & 72.3 & 75.2 & 75.0 & 78.6 \\
 & S$^2$I-MI-FGSM(ours) & \textbf{89.8} & \textbf{89.0} & 98.4* & \textbf{84.9} & \textbf{86.0} & \textbf{84.3} & \textbf{88.7} \\
 \hline
\multirow{9}{*}{Res-152} & MI-FGSM & 54.7 & 50.1 & 45.5 & 99.4* & 84.0 & 86.5 & 70.0 \\
 & DI-FGSM & 57.3 & 51.5 & 47.2 & 99.3* & 83.1 & 85.1 & 70.6 \\
 & PI-FGSM & 63.2 & 55.1 & 47.8 & \textbf{99.7*} & 82.8 & 84.8 & 72.2 \\
 & S$^2$I-FGSM(ours) & \textbf{66.8} & \textbf{62.8} & \textbf{57.4} & \textbf{99.7*} & \textbf{92.8} & \textbf{94.4} & \textbf{79.0} \\
 \cline{2-9}
 & SI-NI-FGSM & 75.3 & 72.9 & 70.2 & \textbf{99.7*} & 94.5 & 94.9 & 84.6 \\
 & VT-MI-FGSM & 73.7 & 69.4 & 66.4 & 99.5* & 93.1 & 93.8 & 82.7 \\
 & FI-MI-FGSM & 83.7 & 82.1 & 78.6 & 99.4* & 93.6 & 94.2 & 88.6 \\
 & S$^2$I-MI-FGSM(ours) & \textbf{88.1} & \textbf{86.9} & \textbf{86.3} & \textbf{99.7*} & \textbf{97.5} & \textbf{97.6} & \textbf{92.7}\\
 \hline
\end{tabular}}
\label{tab:normal}
\end{table*}

\begin{table*}[h]
\centering
\caption{The attack success rates (\%) on nine defenses. The adversarial examples are crafted via Inc-v3, Inc-v4, IncRes-v2 and Res-152,  respectively.}
\resizebox{0.95\linewidth}{!}{
\begin{tabular}{c|c|p{1.4cm}<{\centering}|p{1.4cm}<{\centering}|p{1.2cm}<{\centering}|p{1cm}<{\centering}|p{1cm}<{\centering}|c|p{1cm}<{\centering}|p{1cm}<{\centering}|p{1cm}<{\centering}|c}
\hline
\multirow{2}{*}{Model} & \multirow{2}{*}{Attack} & \multirow{2}{*}{Inc-v3$_{ens3}$} & \multirow{2}{*}{Inc-v3$_{ens4}$} & IncRes-v2$_{ens}$ & \multirow{2}{*}{HGD} & \multirow{2}{*}{R\&P} & \multirow{2}{*}{NIPS-r3} & \multirow{2}{*}{JPEG} & \multirow{2}{*}{RS} & \multirow{2}{*}{NRP} & \multirow{2}{*}{AVG.} \\
\hline
\hline
\multirow{8}{*}{Inc-v3} 
% & DIM & 32.0 & 31.9 & 17.0 & 17.2 & 17.3 & 24.5 & 51.9 & 46.8 & 20.7 & 28.8 \\
 & TI-DIM & 43.2 & 42.1 & 27.9 & 36.0 & 30.2 & 37.4 & 56.7 & 55.8 & 22.0 & 39.0 \\ 
 & PI-TI-DI-FGSM & 43.5 & 46.3 & 35.3 & 33.9 & 35.2 & 39.9 & 47.6 & \textbf{74.9} & 37.0 & 43.7 \\
 & SI-NI-TI-DIM & 55.0 & 53.0 & 36.5 & 37.0 & 37.9 & 48.5 & 72.3 & 55.2 & 32.7 & 47.6 \\
 & VT-TI-DIM & 61.3 & 60.4 & 46.6 & 53.9 & 47.8 & 53.3 & 68.3 & 62.4 & 36.1 & 54.5 \\
 & FI-TI-DIM & 61.8 & 59.6 & 49.2 & 51.7 & 48.3 & 55.0 & 71.3 & 64.5 & 38.0 & 55.5 \\
 & Admix-TI-DIM & 75.3 & 72.1 & 56.7 & 65.8 & 59.8 & 66.0 & 83.7 & 70.5 & 45.3 & 66.1 \\
 & S$^2$I-TI-DIM (ours)& 81.5 & 81.2 & 69.8 & 77.8 & 70.1 & 77.2 & 86.7 & 71.8 & \textbf{56.0} & 74.7 \\
 & S$^2$I-SI-DIM (ours)& 83.8 & 81.8 & 64.8 & 71.1 & 68.9 & 77.4& \textbf{91.8}& 72.6 & 52.3 & 73.8 \\
 & S$^2$I-SI-TI-DIM (ours)& \textbf{88.6} & \textbf{87.8} & \textbf{77.9} & \textbf{81.1} & \textbf{77.6} & \textbf{83.3}& 91.3 & 71.0 & 55.1 & \textbf{79.3} \\
 \hline
\multirow{8}{*}{Inc-v4} 
% & DIM & 27.6 & 26.3 & 15.2 & 15.6 & 16.3 & 20.8 & 47.5 & 46.3 & 16.6 & 25.8 \\
 & TI-DIM & 38.4 & 38.1 & 27.7 & 33.7 & 29.5 & 33.0 & 51.2 &55.0 & 19.0 & 36.2 \\
 & PI-TI-DI-FGSM & 42.3 & 43.8 & 32.5 & 33.0 & 33.9 & 36.7 & 46.0 & 74.8 & 32.3 & 41.7 \\
 & SI-NI-TI-DIM & 60.2 & 56.9 & 43.8 & 46.0 & 46.5 & 52.7 & 73.7 & 56.3 & 32.5 & 52.1 \\
 & VT-TI-DIM & 57.7 & 57.2 & 46.9 & 55.1 & 48.9 & 50.4 & 63.3 & 59.1 & 34.9 & 52.6 \\
 & FI-TI-DIM & 61.0 & 58.4 & 50.6 & 53.6 & 51.7 & 55.1 & 67.7 & 62.6 & 38.6 & 55.5 \\
 & Admix-TI-DIM & 77.3 & 74.1 & 63.8 & 73.4 & 67.1 & 71.4 & 82.6 & 67.2 & 48.0 & 69.4 \\
 & S$^2$I-TI-DIM (ours)& 78.7 & 78.0 & 69.9 & 76.6 & 71.9 & 77.1 & 83.5 & 73.4 & 55.0 & 73.8 \\
 & S$^2$I-SI-DIM (ours)& 86.0 & 83.7 & 72.4 & 78.4 & 76.8 & 81.7 & \textbf{91.2} & 73.9 & \textbf{60.9} & 78.3 \\
  & S$^2$I-SI-TI-DIM (ours)& \textbf{88.7} & \textbf{87.7} & \textbf{81.7} & \textbf{86.1} & \textbf{83.5} & \textbf{86.3}& 90.8 & \textbf{75.0} & 59.6 & \textbf{82.2} \\
 \hline
\multirow{8}{*}{IncRes-v2} 
% & DIM & 33.7 & 31.0 & 22.2 & 24.7 & 23.2 & 27.2 & 52.3 & 48.4 & 19.1 & 31.3 \\
 & TI-DIM & 48.0 & 43.6 & 38.9 & 43.9 & 40.5 & 43.2 & 57.3 &57.3 & 24.7 & 44.2 \\
 & PI-TI-DI-FGSM & 49.7 & 51.1 & 46.0 & 40.1 & 45.9 & 47.8 & 50.6 &78.0 & 41.0 & 50.0 \\
 & SI-NI-TI-DIM & 71.8 & 62.8 & 55.6 & 53.2 & 59.6 & 64.7 & 82.0 & 60.6 & 41.0 & 61.3 \\
 & VT-TI-DIM & 65.9 & 60.1 & 58.2 & 60.3 & 57.6 & 60.1 & 70.1 & 61.2 & 36.9 & 58.9 \\
 & FI-TI-DIM & 58.1 & 54.4 & 53.5 & 52.6 & 52.2 & 56.8 & 64.2 & 64.4 & 39.8 & 55.1 \\
 & Admix-TI-DIM & 85.3 & 82.0 & 79.5 & 82.4 & 79.6 & 82.4 & 85.9 & 74.2 & 59.7 & 79.0 \\
 & S$^2$I-TI-DIM (ours)& 82.6 & 79.9 & 79.2 & 79.5 & 79.3 & 81.2 & 86.1 &74.2 & 61.6 & 78.2 \\
 & S$^2$I-SI-DIM (ours)& 90.3 & 88.6 & 83.7 & 86.6 & 84.1 & 86.9 & 92.0 & 75.5 & 69.0 & 84.1 \\
 & S$^2$I-SI-TI-DIM (ours)& \textbf{92.1} & \textbf{91.0} & \textbf{90.6} & \textbf{90.8} & \textbf{89.2} & \textbf{90.9}& \textbf{93.3} & \textbf{79.2} & \textbf{73.4} & \textbf{87.8} \\
 \hline
\multirow{8}{*}{Res-152} 
% & DIM & 41.5 & 39.7 & 27.4 & 34.6 & 28.9 & 34.6 & 60.8 & 52.5 & 24.1 & 38.2 \\
 & TI-DIM & 55.1 & 52.3 & 42.5 & 55.6 & 46.5 & 52.3 & 64.9 & 61.2 & 32.2 & 51.4 \\
 & PI-TI-DI-FGSM & 54.3 & 56.2 & 45.3 & 43.7 & 46.2 & 48.9 & 55.2 & 78.1 & 47.7 & 52.8 \\
 & SI-NI-TI-DIM & 68.6 & 64.0 & 52.4 & 58.9 & 56.8 & 64.2 & 80.1 & 67.5 & 42.3 & 61.6 \\
 & VT-TI-DIM & 64.3 & 61.4 & 54.9 & 60.7 & 54.8 & 59.4 & 69.3 & 67.9 & 41.2 & 59.3 \\
 & FI-TI-DIM & 70.1 & 66.0 & 59.5 & 63.9 & 60.8 & 66.0 & 77.5 & 71.0 & 47.2 & 64.7 \\
 & Admix-TI-DIM & 83.7 & 81.4 & 73.7 & 81.2 & 77.0 & 80.1 & 87.8 & 75.0 & 59.5 & 77.7 \\
 & S$^2$I-TI-DIM (ours)& 86.6 & 83.9 & 79.0 & 85.3 & 81.8 & 85.5 & 90.6 & 80.9 & 66.1 & 82.2 \\
 & S$^2$I-SI-DIM (ours) & 89.3 & 84.4 & 77.9 & 86.6 & 82.7 & 86.3 & \textbf{92.8} & 76.4 & 65.9 & 82.5\\
 & S$^2$I-SI-TI-DIM (ours)& \textbf{92.5} & \textbf{88.6} & \textbf{85.3} & \textbf{88.6} & \textbf{87.8} & \textbf{89.8}& 92.4 & \textbf{83.6} & \textbf{72.0} & \textbf{86.7} \\
 \hline
\end{tabular}}

\label{tab:defense}
\end{table*}

\begin{table*}[h]
\centering
\caption{The attack success rates (\%) on nine defenses. The adversarial examples are crafted via an ensemble of Inc-v3, Inc-v4, IncRes-v2 and Res-152 and the weight for each model is 1/4.}
\resizebox{0.88\linewidth}{!}{
\begin{tabular}{c|p{1.4cm}<{\centering}|p{1.4cm}<{\centering}|p{1.2cm}<{\centering}|p{1cm}<{\centering}|p{1cm}<{\centering}|c|p{1cm}<{\centering}|p{1cm}<{\centering}|p{1cm}<{\centering}|c}
\hline
\multirow{2}{*}{Attack}  & \multirow{2}{*}{Inc-v3$_{ens3}$} & \multirow{2}{*}{Inc-v3$_{ens4}$} & IncRes-v2$_{ens}$ & \multirow{2}{*}{HGD} & \multirow{2}{*}{R\&P} & \multirow{2}{*}{NIPS-r3} & \multirow{2}{*}{JPEG} & \multirow{2}{*}{RS} & \multirow{2}{*}{NRP} & \multirow{2}{*}{AVG.} \\
\hline
\hline
% \hline
% DIM & 67.9 & 63.2 & 47.8 & 61.4 & 52.6 & 62.8 &85.7 &58.0 & 32.4 & 59.1 \\
% \hline
TI-DIM & 79.2 & 75.3 & 69.3 & 80.4 & 73.9 & 76.7 &87.5& 68.3 & 43.1 & 72.6 \\
% \hline
PI-TI-DI-FGSM & 75.0 & 76.0 & 67.7 & 69.5 & 68.0 & 72.6&77.8 & 83.4 & 60.8 & 72.3 \\
% \hline
SI-NI-TI-DIM & 90.2 & 87.9 & 80.0 & 83.2 & 83.5 & 87.8 & 94.3 & 81.4 & 59.2 & 83.1\\
% \hline
VT-TI-DIM & 85.0 & 82.3 & 78.3 & 83.9 & 79.4 & 81.9 &88.5 &74.5 & 59.7 & 79.3 \\
% \hline
FI-TI-DIM & 83.1 & 83.6 & 74.6 & 84.9 & 76.5 & 78.6 & 90.2&72.2 & 61.2 & 78.3 \\
Admix-TI-DIM & 93.9 & 92.9 & 90.3 & 94.0 & 91.3 & 92.0 &95.6 &82.4 & 76.0 & 89.8 \\
S$^2$I-TI-DIM (ours) & 94.6 & 94.3 & 92.5 & 94.3 & 93.1 & 94.3 &95.8& 87.4 & 83.5 & 92.2 \\
% \hline
S$^2$I-SI-DIM (ours) & 96.5 & 96.3 & 94.2 & 95.8 & 94.9 & 96.0 & \textbf{97.4} & 88.2 & 87.3 & 94.1 \\
S$^2$I-SI-TI-DIM (ours)& \textbf{96.7} & \textbf{96.7} & \textbf{95.2} & \textbf{96.3} & \textbf{95.7} & \textbf{96.5}& 96.9 & \textbf{92.2} & \textbf{92.2} & \textbf{95.4} \\
\hline
\end{tabular}}

\label{tab:ensemble}
\end{table*}

\subsection{Attack Normally Trained Models}
In this section, we investigate the vulnerability of normally trained models. 
We first compare S$^2$I-FGSM with MI-FGSM \cite{mifgsm}, DI-FGSM \cite{difgsm}, PI-FGSM \cite{pifgsm} to verify the effectiveness of our method in Table \ref{tab:normal}. 
A first glance shows that S$^2$I-FGSM consistently surpasses well-known baseline attacks on all black-box models. For example, when attacking against Inc-v3, MI-FGSM, DI-FGSM and PI-FGSM only successfully transfer 47.2\%, 38.2\% and 49.6\% adversarial examples to IncRes-v2, while our S$^2$I-FGSM can achieve a much higher success rate of \textbf{58.9\%}. This convincingly validates the high effectiveness of our proposed method against normally trained models.

Besides, we also report the results for methods with the momentum term~\cite{mifgsm}. As displayed in Table \ref{tab:normal}, the performance gap between our proposed method and state-of-the-art approaches is still large. Notably, adversarial examples crafted by our proposed S$^2$I-MI-FGSM are capable of getting \textbf{88.8\%} success rate on average, which outperforms SI-NI-FGSM, VT-MI-FGSM and FI-MI-FGSM by 7.3\%, 12.2\% and 6.3\%, respectively.
% when attacking with our proposed S$^2$I-MI-FGSM, resulting adversarial examples are capable of getting \textbf{88.8\%} success rate on average. 
% In contrast, SI-NI-FGSM, VT-MI-FGSM and FI-MI-FGSM only obtain 81.6\%, 76.6\% and 82.5\%, respectively.
This also demonstrates that the combination of our method and existing attacks can significantly enhance the transferability of adversarial examples.

\subsection{Attack Defense Models}
Although many attack methods can easily fool normally trained models, they may fail in attacking models with the defense mechanism. To further verify the superiority of our method, we conduct a series of experiments against defense models. Given that the vanilla versions of attacks are less effective for defense models, we consider the stronger DIM, TI-DIM, PI-TI-DI-FGSM, SI-NI-TI-DIM, VT-TI-DIM, FI-TI-DIM and Admix-TI-DIM as competitors to our proposed S$^2$I-TI-DIM, S$^2$I-SI-DIM and S$^2$I-SI-TI-DIM.

{\bfseries{Single-Model Attacks.}} We first investigate the transferability of adversarial examples crafted via a single substitute model. From the results of Table~\ref{tab:defense}, we can observe that our algorithm can significantly boost existing attacks. For example, suppose we generate adversarial examples via Inc-v3, TI-DIM only achieves an average success rate of 39.0\% on the nine defense models, while our proposed S$^2$I-TI-DIM can yield about 2$\times$ transferability, \textit{i.e.}, outperforms TI-DIM by \textbf{35.7\%}. This demonstrates the remarkable effectiveness of our proposed method against defense models.

{\bfseries{Ensemble-based Attacks.}} We also report the results for attacking an ensemble of models simultaneously~\cite{liu2017delving} to demonstrate the effectiveness of our proposed method. In particular, the adversarial examples are crafted via an ensemble of Inc-v3, Inc-v4, IncRes-v2 and Res-152. Similar to the results of Table~\ref{tab:defense}, our S$^2$I-SI-TI-DIM displayed in Table~\ref{tab:ensemble} still consistently surpass state-of-the-art approaches.
Remarkably, S$^2$I-SI-TI-DIM is capable of obtaining \textbf{95.4\%} success rate on average, which outperforms SI-NI-TI-DIM, VT-TI-DIM, FI-TI-DIM and Admix-TI-DIM by 23.1\%, 12.4\%, 16.1\%, 17.1\% and 5.6\%, respectively.
This also reveals that current defense mechanisms are still vulnerable to well-design adversarial examples and far from the need of real security.

\begin{figure}[t]
    \centering
    \includegraphics[width=9.25cm]{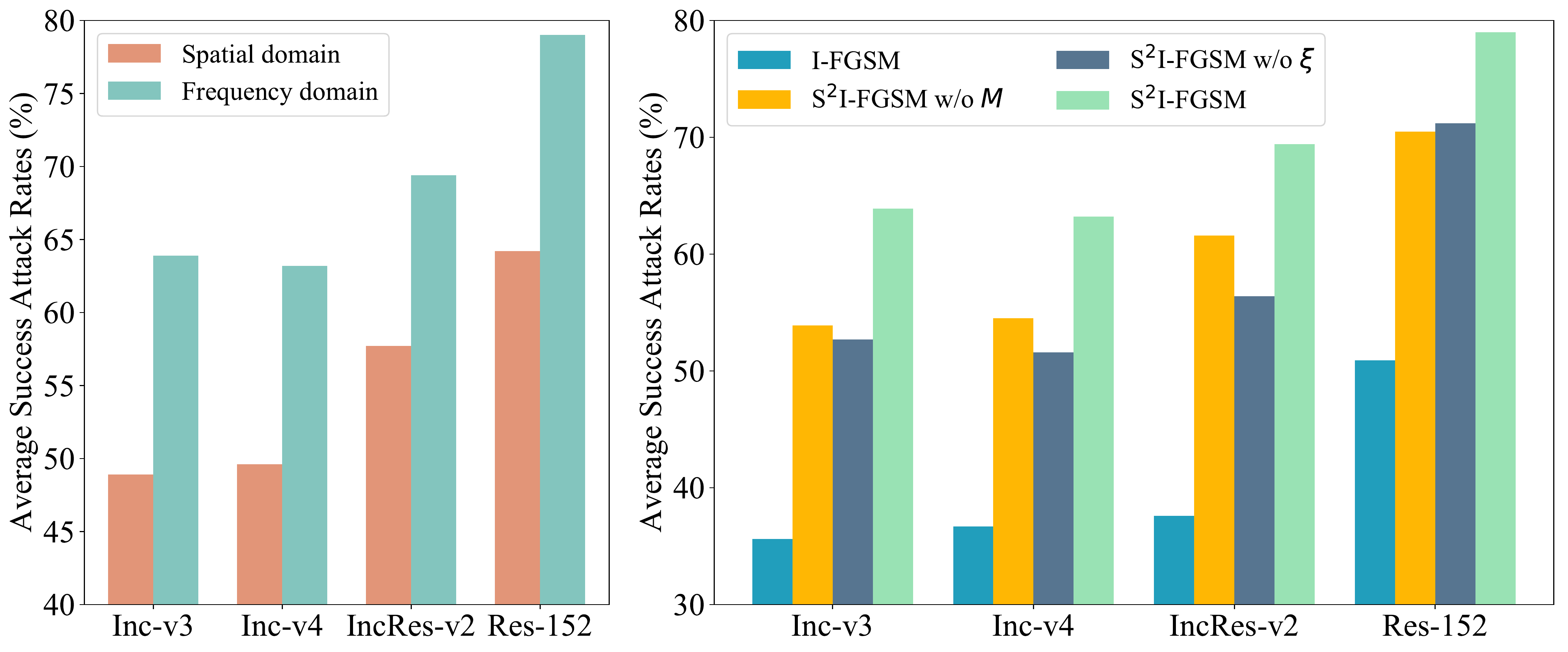}
    \caption{The average attack success rates (\%) on six normally trained models (introduced in Sec.~\ref{setup}). The adversarial examples are crafted via Inc-v3, Inc-v4, IncRes-v2 and Res-152, respectively. \textbf{Left}: Effect of frequency domain transformation. \textbf{Right}: Effect of \bm{$\xi$} and \bm{$M$}.}
    \label{fig:svsf}
\end{figure}

\subsection{Ablation Study}

In this section, we analyze the impact of different aspects of our method:

% {\bfseries{Block dct vs. Overall dct.}} Here we evaluate the impact of different dct strategies on attack performance. We adopt two dct strategies: a) block dct: we first split the original images into blocks with size N × N, then apply dct for each block. b) overall dct: apply dct for a full image. We compare attack success rate of the two strategies. From the result in Table~\ref{table4}, despite the increased computational overhead of the overall dct, it has a much higher attack success rate than block dct. We hold that the spectrum transformation expects to change the frequency information of the full image, therefore overall dct is a better choice.

% \begin{table*}[h]
% \centering
% \caption{The attack success rates (\%) of black-box attacks against  six normally trained models. The adversarial examples are crafted via Inc-v3.}
% \begin{tabular}{c|c|c|c|c|c|c|c}
% \hline
% Attack & \multicolumn{1}{l}{Inc-v3} & \multicolumn{1}{l}{Inc-v4} & \multicolumn{1}{l}{IncRes-v2} & \multicolumn{1}{l}{Res-152} & \multicolumn{1}{l}{Res-50} & \multicolumn{1}{l}{Res-101} & Average \\
% \hline
% \hline
% Block & 80.0 & 24.4 & 20.3 & 22.7 & 28.2 & 21.9 & 32.9 \\
% Overall & \textbf{99.7} & \textbf{65.0} & \textbf{58.9} & \textbf{50.3} & \textbf{56.2} & \textbf{53.3} & \textbf{63.9}\\
% \hline
% \end{tabular}
% \label{table4}
% \end{table*}

{\bfseries{Frequency domain vs. Spatial domain.}} For our proposed S$^2$I-FGSM, transformation is applied in the frequency domain. To verify that frequency domain transformation (\textit{i.e.}, our spectrum transformation) is more potent in narrowing the gap between models than spatial domain transformation (\textit{i.e.}, remove the DCT/IDCT in spectrum transformation), we conduction an ablation study. 
% Specifically, we compare the results of frequency domain attack and spatial domain attack (\textit{i.e.}, remove the DCT/IDCT in the spectrum transformation). 
As depicted in Figure~\ref{fig:svsf} (left), regardless of what substitute models are attacked, the transferability of adversarial examples crafted based on frequency domain transformation is consistently higher than that of spatial domain transformation. Notably, when attacking against Inc-v3, the attack based on frequency domain transformation (\textit{i.e.}, S$^2$I-FGSM) outperforms the attack based on spatial domain transformation by a large margin of \textbf{15.0\%}. This convincingly validates that frequency domain can capture more essential differences among models, thus yielding more diverse substitute models than spatial domain.

{\bfseries{Effect of \bm{$\xi$} and \bm{$M$}.}} To analyze the effect of each random variant (\textit{i.e.}, \bm{$\xi$} and \bm{$M$}) in our spectrum transformation, we conduct the experiment in Figure~\ref{fig:svsf} (right). From this result, we observe that both \bm{$\xi$} and \bm{$M$} are useful for enhancing the transferability of adversarial examples. It is because both of them can manipulate the spectrum saliency map to a certain extent, albeit from different aspects of implementation. Therefore, by leveraging them simultaneously, our proposed spectrum transformation can simulate a more diverse substitute model, thus significantly boosting attacks.

\begin{figure}[h]
    \centering
    \includegraphics[width=9cm]{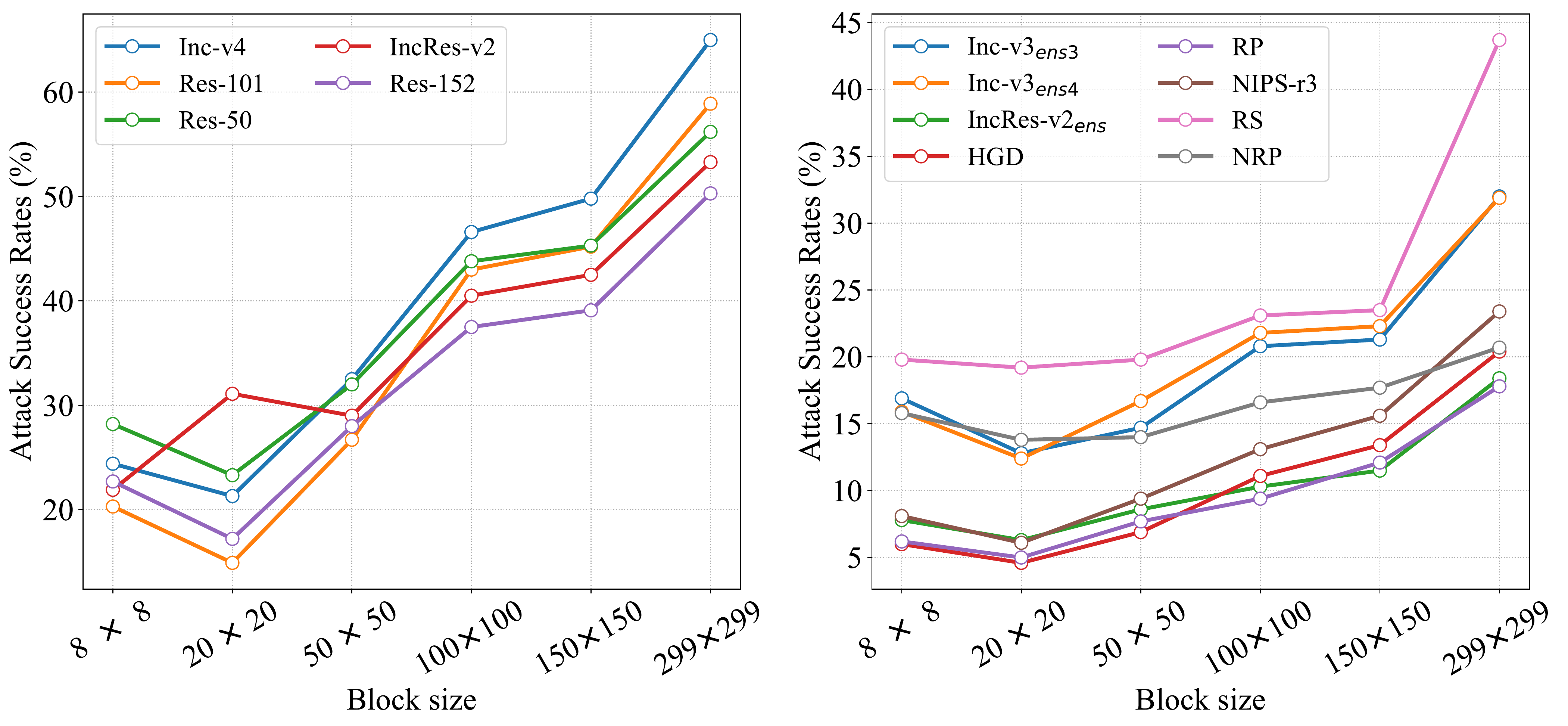}
    \caption{The attack success rates (\%) of S$^2$I-FGSM on normally trained models (\textbf{Left}) and defense models (\textbf{Right}) w.r.t. the block size of DCT/IDCT. Adversarial examples are generated via Inc-v3.}
    \label{fig:block}
\end{figure}

{\bfseries{On the block size of DCT/IDCT.}}
Previous works~\cite{advdrop,projection} usually started by splitting images into small blocks with size $n\times n$ and then apply DCT/IDCT. However, it is not clear that this paradigm is appropriate for our approach. Therefore, in this part, we investigate the impact of block size on the transferability. Specifically, we tune the block size from $8\times 8$ to $299\times299$ (full image size) and report the attack success rates of S$^2$I-FGSM in Figure~\ref{fig:block}. From this result, we observe that larger blocks are more suited to our approach. Particularly, the attack success rates reach peak when the size of the block is the same as the full image size. Therefore, in our experiment, we do not split the image beforehand and directly apply DCT/IDCT on the full image to get its spectrum (we also provide time analysis of DCT/IDCT in supplementary Sec. C). 

\begin{figure*}[h]
    \centering
    \includegraphics[width=12cm]{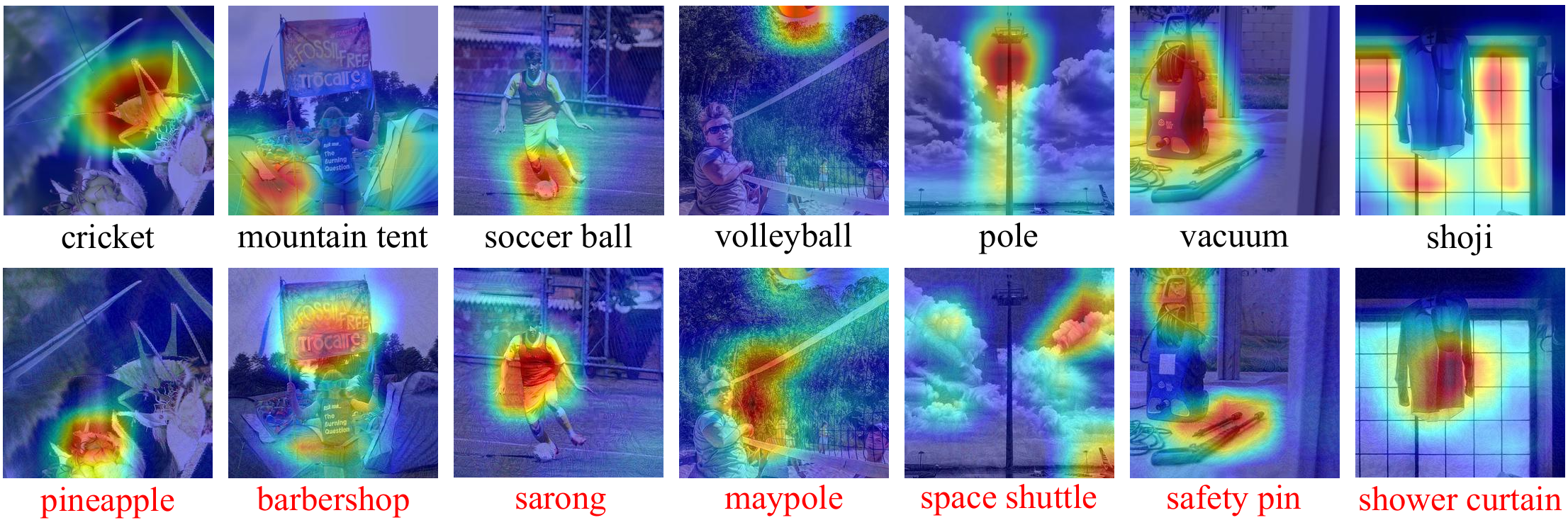}
    \caption{Visualization for attention shift. We apply Grad-CAM~\cite{grad_cam} for Res-152~\cite{res152} to visualize attention maps of clean (1st row) and adversarial images (2nd row). Adversarial examples are crafted via Inc-v3 by our S$^2$I-FGSM. The result demonstrates that our adversarial examples are capable of shifting model's attention.}
    \label{fig:attention}
\end{figure*}

{\bfseries{Attention shift.}}
To better understand the effectiveness of our attack, we apply Grad-CAM~\cite{grad_cam} to compare attention maps of clean images with those of adversarial examples. As illustrated in Figure~\ref{fig:attention}, our proposed method can effectively shift the model's attention from the key object to other mismatched regions.
Consequently, the victim model inevitably captures other irrelevant features, thus leading to misclassification.

\section{Conclusion}
In this paper, we propose a Spectrum Simulation Attack to boost adversarial attacks from a frequency domain perspective. Our work gives a novel insight into model augmentation, which narrows the gap between the substitute model and victim model by a set of spectrum transformation images. We also conduct a detailed ablation study to clearly illustrate the effect of each component. 
% Our work not only significantly improves the attack success rate, but also provides a way of model augmentation based on the frequency domain.
% crafting adversarial examples in frequency domain. 
Compared with traditional model augmentation attacks in spatial domain, extensive experiments demonstrate the significant effectiveness of our method, which outperforms state-of-the-art transfer-based attacks by a large margin. 
% Besides, our method can be integrated with existing gradient-based attacks to further enhance the transferability of adversarial examples.

\section{Acknowledge}
This work is supported by the National Natural Science Foundation of China (Grant No. 62122018, No. 61772116, No. 61872064, No. U20B2063).
% Notably, our attacks fool nine state-of-the-art defense models at a \textbf{94.1\%} success rate on average. 

%Our method can serve as a benchmark for evaluating the robustness of various models.
%, and the results strongly support our claims. 

% \textbf{Broader Impact \& Potential Harms}. Our proposed method can serve as a benchmark for evaluating the robustness of various models. However, it also may raise new concerns about AI safety, \textit{e.g.}, unscrupulous people may use our method to undermine real-world applications. To address this issue, adversarial training with adversarial examples crafted by our method can be helpful.

%frequency domain attack outperforms spatial domain attack and spectrum transformation boosts the attack. 
%In addition, we also propose a method to reduce generating time without degrading attack performance.
% In the future, we may expand the current approach to explore the weakness of defense models.
%Our work not only significantly improve the attack success rate, but also provide a way of crafting adversarial examples in frequency domain, therefore future work can follow this way to explore the weakness of DNN networks.

\clearpage
\bibliographystyle{splncs04}
\bibliography{egbib}

\clearpage

\appendix

\section{Proof}
\label{supp:A}

\noindent{\bfseries{Proposition 1}}. \textit{Our proposed spectrum transformation can generate diverse spectrum saliency maps and thus simulate diverse substitute models.}

\noindent{\textit{Proof}}. According to \textit{Lagrange's mean value theorem}:
\begin{equation}
    \frac{\partial {J(\bm{x_1},y;\phi)}}{\partial \bm{x_1}}=\frac{\partial {J(\bm{x_2},y;\phi)}}{\partial \bm{x_2}}+\bm{K},
    \label{eq1}
\end{equation}
where $\bm{K} = \frac{\partial ^{2} {J(\bm{\zeta},y;\phi)}}{\partial \bm{\zeta}^{2}} (\bm{x_1} - \bm{x_2}), \bm{\zeta} \in \left[\bm{x_2}, \bm{x_1}\right] $.

Without spectrum transformation function $\mathcal{T}(\cdot)$, spectrum saliency map:
\begin{equation}
    \bm{S}_\phi =  \frac{\partial {J(\mathcal{D_I}(\mathcal{D}(\bm{x})),y;\phi)}}{\partial \mathcal{D}(\bm{x})}, 
\end{equation}
after applying our proposed spectrum transformation function $\mathcal{T}(\cdot)$, the resulting spectrum saliency map:
\begin{equation}
    \bm{S}_\phi' = \frac{\partial {J(\mathcal{T}(\bm{x}),y;\phi)}}{\partial \mathcal{D}(\bm{x})}, 
\end{equation}
where $\mathcal{T}(\bm{x}) = \mathcal{D_I}((\mathcal{D}(\bm{x})+\mathcal{D}(\bm{\xi})) \odot \bm{M})$

Let $\bm{D_1}$ denotes $\frac{\partial {J(\mathcal{D_I}(\mathcal{D}(\bm{x})),y;\phi)}}{\partial \mathcal{D_I}(\mathcal{D}(\bm{x}))}$ and $\bm{D_2}$ denotes $\frac{\partial {\mathcal{D_I}(\mathcal{D}(\bm{x}))}}{\partial \mathcal{D}(\bm{x})}$, then $\bm{S}_\phi = \bm{D_1} \bm{D_2}$ (according to chain rule).
After applying $\mathcal{T}(\cdot)$ to $\bm{x}$, resulting spectrum saliency map $\bm{S}_\phi'$ can be expressed as:
\begin{equation}
    \bm{S}_\phi' = \bm{D_1}' \bm{D_2}' \odot \bm{M},
\end{equation}
where 
\begin{align}
    \bm{D_1}'&=\frac{\partial {J(\mathcal{D_I}(\mathcal{D}(\bm{x}+\bm{\xi})\odot \bm{M}),y;\phi)}}{\partial \mathcal{D_I}(\mathcal{D}(\bm{x}+\bm{\xi})\odot \bm{M})},\\
    \bm{D_2}'&=\frac{\partial {\mathcal{D_I}(\mathcal{D}(\bm{x}+\bm{\xi})\odot \bm{M})}}{\partial \mathcal(\mathcal{D}(\bm{x}+\bm{\xi})\odot \bm{M}))}.
\end{align}
Based on Eq.~\ref{eq1}, we can formally formulate $\bm{S}_\phi'$ to be:
\begin{equation}
\begin{aligned}
\label{eq10}
      \bm{S}_\phi' &= (\bm{D_1}+\bm{K_1})(\bm{D_2}+\bm{K_2})\odot\bm{M},\\
                &= (\bm{S}_\phi+\bm{K'})\odot\bm{M},  
\end{aligned}
\end{equation}
where $\bm{K_1}$ and $\bm{K_2}$ are two specific matrices, and $\bm{K'}=\bm{D_1} \bm{K_2}+\bm{D_2} \bm{K_1}+\bm{K_1} \bm{K_2}$. Eq.~\ref{eq10} clearly demonstrates that our proposed transformation $\mathcal{T}(\cdot)$ is capable of simulating a different spectrum saliency map.

\section{On the Hyper-Parameters Settings}
\label{supp:B}
\label{hyper}
We first study the influence of the hyper-parameters(\textit{i.e.}, standard deviation (std) $\sigma$ of noise \bm{$\xi$}, tuning factor $\rho$ of matrix \bm{$M$}, number $N$ of spectrum transformations) for the proposed Spectrum Simulation Attack method.

% \subsection{On the Block Size of DCT}
% In previous works~\cite{advdrop,projection}, DCT is usually applied on small patches of images which splits the original images into blocks with size n × n, then apply DCT on each block. Here, we study the impact of different block size from 8 to 299 (the full image size). 
% In Figure~\ref{fig:block},  we report the attack success rates of S$^2$I-FGSM for different block size. Adversarial examples are crafted via Inc-v3. When block size exceeds 50, the attack success rates increase significantly with the increase of block size. Particularly, when block size is 299 which is the original image size, the attack success rates reach peak. Therefore, in our experiment, we don't split images into small patches and apply DCT on the full image.
% % When $\sigma = 16$, normally trained models reach the peak and defense models also get higher results.  Then the attack success rates will decrease for normally trained models. To achieve the transferability trade-off on the normally trained models and the defense models, we choose $\sigma= 16$ in our experiments.
% \begin{figure}[h]
%     \centering
%     \includegraphics[width=7.5cm]{images/Ablation images/block.pdf}
%     \caption{The attack success rates (\%) of S$^2$I-FGSM on normally trained w.r.t. the block size of DCT. Adversarial examples are generated via Inc-v3.}
%     \label{fig:block}
% \end{figure}
\subsection{On the Standard Deviation $\sigma$ of Noise $\xi$}
\begin{figure}[h]
    \centering
    \includegraphics[width=8cm]{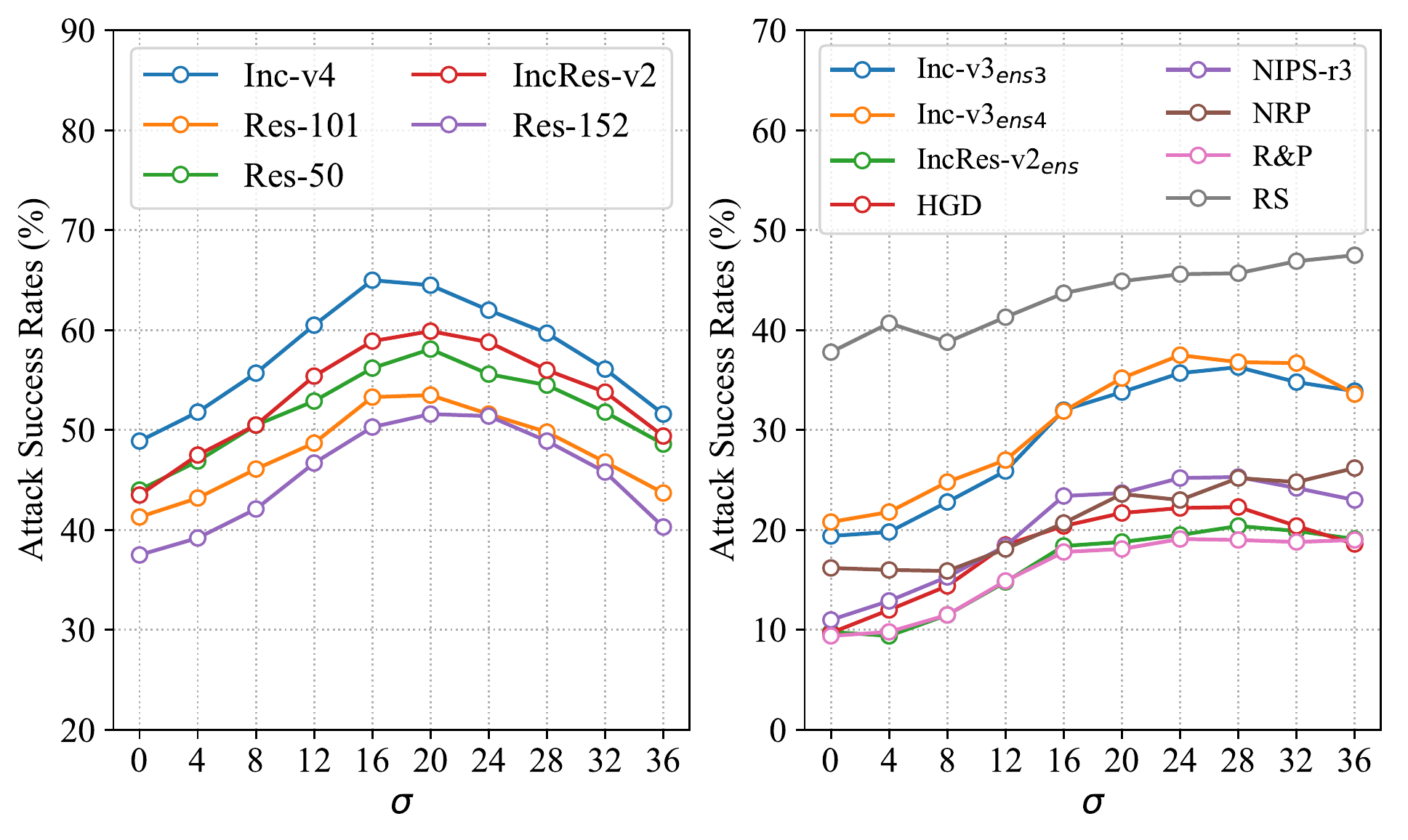}
    \caption{The attack success rates (\%) of S$^2$I-FGSM on normally trained and defense models w.r.t. the std $\sigma$ of $\xi$. Adversarial examples are generated via Inc-v3. \textbf{Left}: The results for fooling normally trained models. \textbf{Right}: The results for fooling defense models.}
    \label{fig:sigma}
\end{figure}

In Figure~\ref{fig:sigma}, we report the attack success rates of S$^2$I-FGSM for different std $\sigma$. Adversarial examples are crafted via Inc-v3 with $N = 20$ and $\rho=0.5$. Particularly, $\sigma = 0$ means no noise is added to the input. A first glance shows that for normally trained models, the attack success rates increase gradually as $\sigma$ increases and then tend to decrease when $\sigma$ exceeds 16. Also when $\sigma = 16$, the defense models can achieve relatively high attack success rates. Therefore, we set $\sigma= 16$ in our paper.
% When $\sigma = 16$, normally trained models reach the peak and defense models also get higher results.  Then the attack success rates will decrease for normally trained models. To achieve the transferability trade-off on the normally trained models and the defense models, we choose $\sigma= 16$ in our experiments.
\subsection{On the Tuning Factor $\rho$ of Matrix \textit{M}}
\begin{figure}
    \centering
    \includegraphics[width=8cm]{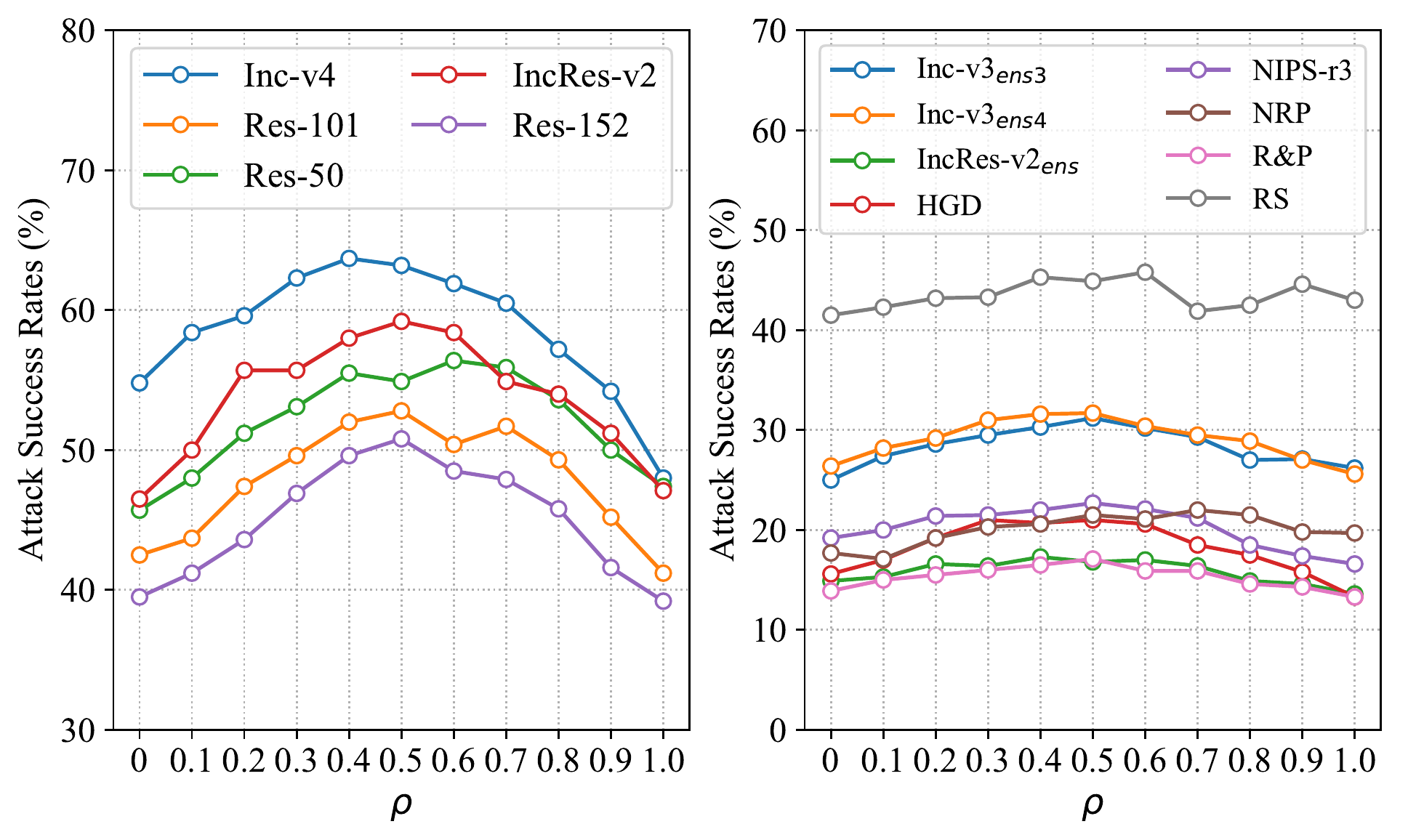}
    \caption{The attack success rates (\%) of S$^2$I-FGSM on normally trained and defense models w.r.t. the tuning factor $\rho$. Adversarial examples are generated via Inc-v3. \textbf{Left}: The results for fooling normally trained models. \textbf{Right}: The results for fooling defense models.}
    \label{fig:rho}
\end{figure}

In this section, we study the effect of tuning factor $\rho$ for our S$^2$I-FGSM in Figure~\ref{fig:rho}. Adversarial examples are crafted via Inc-v3 with $N=20$ and $\sigma=16$. Particularly, $\rho = 0$ means there is no tuning on the spectrum. Similarly, as $\rho$ increases, the degree of spectrum transformation becomes stronger and the attack success rates gradually increase and peak at $\rho=0.5$. If we continue to increase $\rho$ (\textit{i.e.} $\rho > 0.5$), the attack success rates will decrease which may be attributed to the excessive spectrum transformation. To achieve better transferability, we choose $\rho = 0.5$ in our paper.
\subsection{On the Number $N$ of Spectrum Transformations.}
\begin{figure}
    \centering
    \includegraphics[width=8cm]{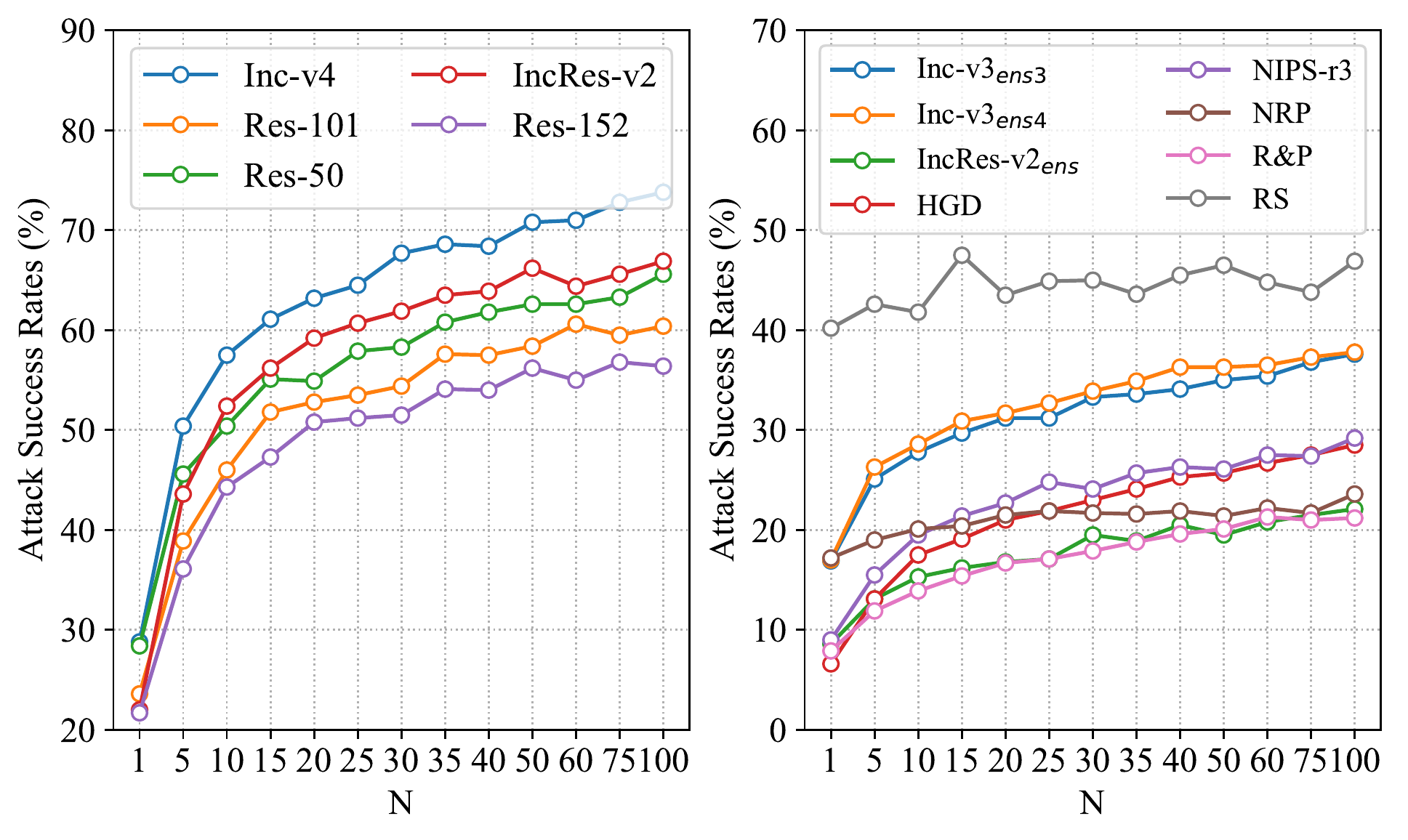}
    \caption{The attack success rates (\%) of S$^2$I-FGSM on normally trained and defense models w.r.t. the number $N$ of spectrum transformations. Adversarial examples are generated via Inc-v3. \textbf{Left}: The results for fooling normally trained models. \textbf{Right}: The results for fooling defense models.}
    \label{fig:number}
\end{figure}

In this section, we study the effect of number $N$ of spectrum transformations for our S$^2$I-FGSM in Figure~\ref{fig:number}. Adversarial examples are crafted via Inc-v3 with $\rho=0.5$ and $\sigma=16$.
As shown in Figure~\ref{fig:number}, when $N= 1$, our method performs only one spectrum transformation and achieves the lowest transferability. As $N$ increases, the transferability of adversarial examples is significantly enhanced at first, and turns to increase slowly after $N$ exceeds 20. It also demonstrates that our spectrum transformation can effectively narrow the gap between the substitute model and victim model.
It is worth noting that larger $N$ implies expensive computational overhead, as we need more forward and backward propagation for gradient computation at each iteration. To balance the transferability and computational overhead, we choose $N=20$ in our paper.

% \section{Result visualization}
\section{Time Analysis of DCT/IDCT}
In our experiments, we directly apply DCT/IDCT on the full image which is a time-consuming operation. Therefore, in this section we analyze the time consumption of DCT/IDCT. In Tab.\ref{tabb} we show the average time of an adversarial example generated by S$^2$I-FGSM and the average time of DCT/IDCT among it. For example, let IncRes-v2 be the substitute model, S$^2$I-FGSM takes an average of 3.78s to produce an adversarial example, of which DCT/IDCT takes up 0.58s (only accounts for 15.3\% of all overheads). The experiment is conducted on RTX 3090 GPUs.

\begin{table}[]

\centering
\caption{The average time (s) of generating an adversarial example on Inc-v3, Inc-v4, IncRes-v2 and Res-152, respectively. The left side of slash indicates the time of DCT/IDCT and right side indicates the time of S$^2$I-FGSM.}
\setlength{\tabcolsep}{5mm}{
\begin{tabular}{l|l|l|l|l}
\hline
 & \multicolumn{1}{c|}{Inc-v3} & \multicolumn{1}{c|}{Inc-v4} & \multicolumn{1}{c|}{IncRes-v2} & \multicolumn{1}{c}{Res-152} \\ \hline
Time & 0.60/1.89 & 0.61/2.85 & 0.58/3.78 & 0.61/3.05 \\ \hline
\end{tabular}}
\label{tabb}
\end{table}

\section{Additional Results}
\label{supp:C}
\subsection{Spatial Domain Transformation Analysis}
\begin{figure}[h]
    \centering
    \includegraphics[width=8cm]{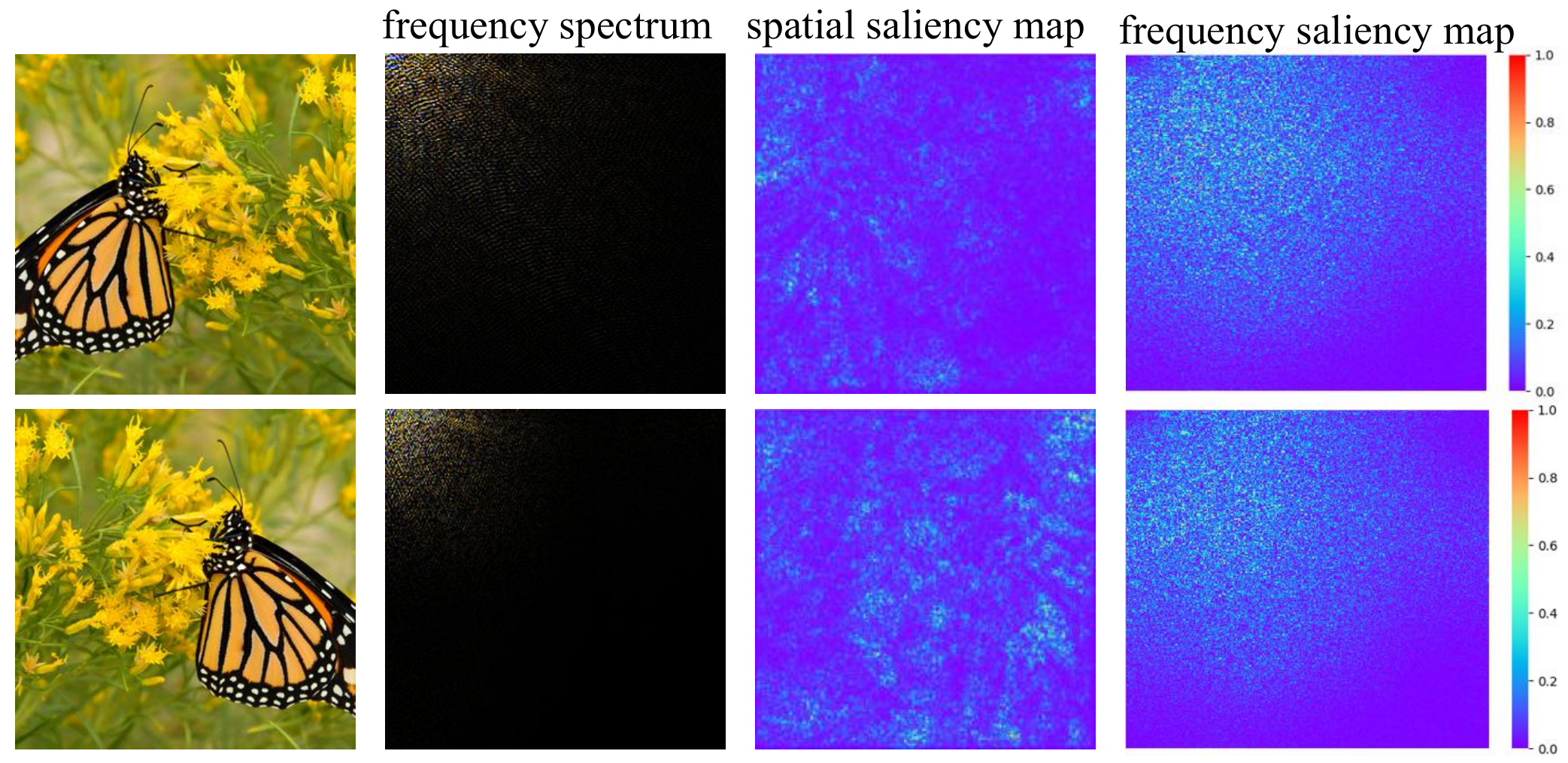}
    \caption{Visualization for frequency spectrum, spatial saliency map, and frequency saliency map. Top raw corresponds to raw image, and bottom row corresponds to spatial domain transformed image. This result demonstrates that analysis on spatial domain is unreliable.}
    \label{fig:s-f}
\end{figure}

In this section, we further validate our point that analysis on spatial domain cannot  well reflect the gap between models. To support our point, we first define spatial saliency map $\bm{\hat{S}_{\phi}}$ as:
\begin{equation}
\begin{aligned}
    \bm{\hat{S}_{\phi}} = \frac{\partial {J(\bm{x}, y;\phi)}}{\partial \bm{x}},
\end{aligned}
\end{equation}
which is similar to our proposed spectrum saliency map \bm{$S_{\phi}$} in Eq. 4. 
Then we flip the image horizontally (spatial domain transformation) and analyze their spatial saliency map and frequency saliency map. As shown in Figure~\ref{fig:s-f}, although spatial saliency maps between raw image and fliped image vary greatly, the changes in frequency spectrum and frequency saliency map (an indicator reflecting the characteristics of models) are small. Thus, analysis on spatial domain is unreliable and can hardly reflect the gap between models.

\subsection{Spectrum Transformation Images}
\begin{figure*}[h]
    \centering
    \includegraphics[width=12cm]{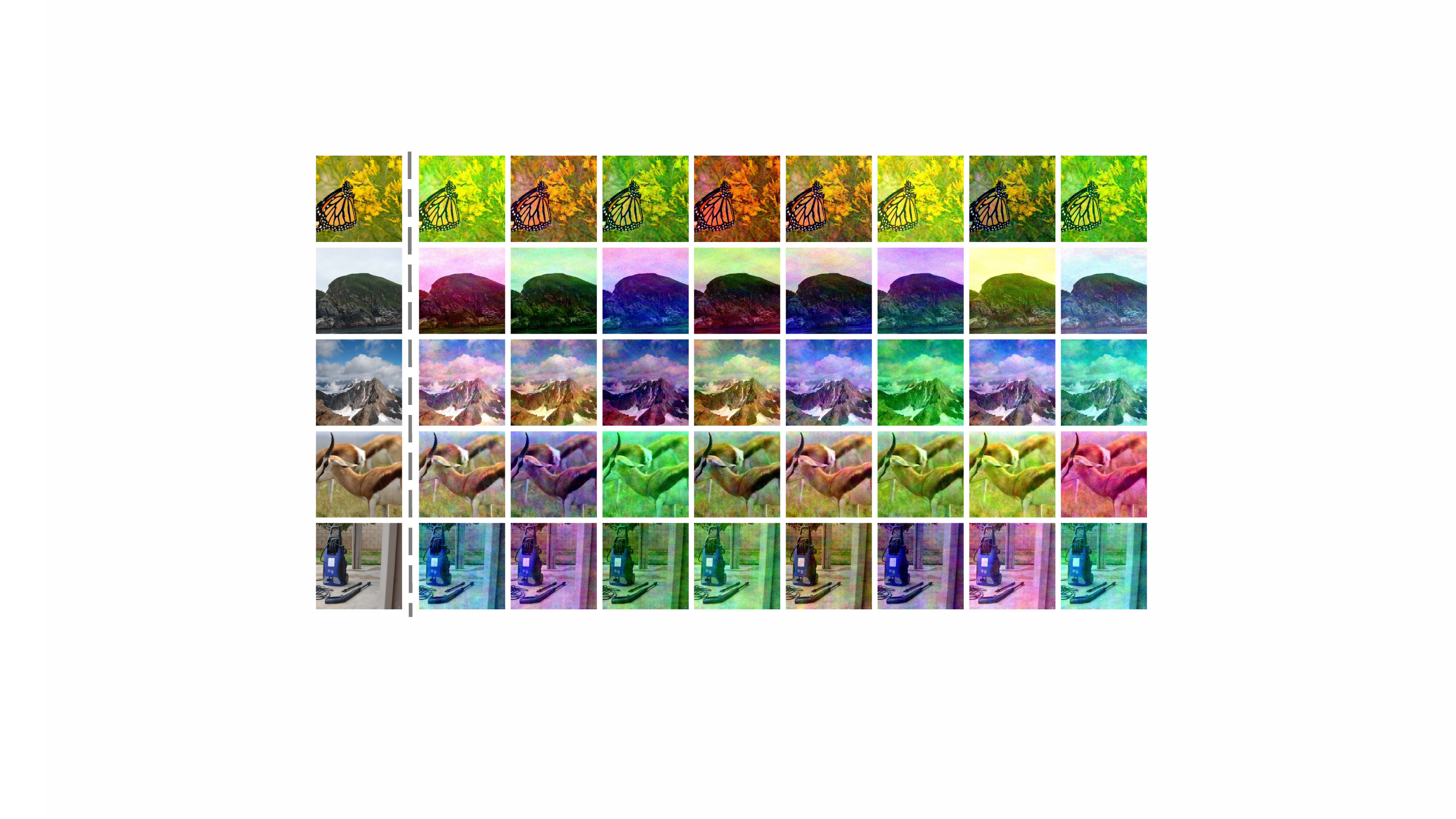}
    \caption{Visualization for the spectrum transformation outputs (right columns) w.r.t. raw input images (left column). This result shows that spectrum transformation just modifies colors of image and does not change its semantic information.}
    \label{fig:transformation}
\end{figure*}

To better understand the process of our method, we visualize the outputs of spectrum transformation. Specifically, we perform several spectrum transformations on input images and show the resulting spectrum transformation outputs in Figure~\ref{fig:transformation}.
This figure shows that spectrum transformation just modifies colors of image and does not change its semantic information.

\end{document}